\documentclass[10pt,twocolumn,letterpaper]{article}

\usepackage{wacv}
\usepackage{times}
\usepackage{epsfig}
\usepackage{graphicx}
\usepackage{tabularx}
\usepackage{amsmath}
\usepackage{amssymb}
\usepackage{soul}
\usepackage{todo}
\usepackage{enumitem}
\usepackage{caption}
\usepackage{subcaption}
\usepackage{float}
\usepackage[accsupp]{axessibility}

\def\wacvPaperID{617} % Enter the WACV Paper ID here

\wacvfinalcopy % *** Uncomment this line for the final submission

\ifwacvfinal
\fi

\ifwacvfinal
\usepackage[breaklinks=true,bookmarks=false]{hyperref}
\else
\usepackage[pagebackref=true,breaklinks=true,colorlinks,bookmarks=false]{hyperref}
\fi

\newcommand{\cU}{\mathcal{U}}
\newcommand{\cV}{\mathcal{V}}
\newcommand{\cT}{\mathcal{T}}
\newcommand{\cB}{\mathcal{B}}
\newcommand{\cS}{\mathcal{S}}
\newcommand{\cF}{\mathcal{F}}
\newcommand{\cG}{\mathcal{G}}

\begin{document}

%%%%%%%%% TITLE
%\title{Attacking a Satellite Imagery Detector with Physical Adversarial Examples}

\title{Physical Adversarial Attacks on an Aerial Imagery Object Detector}

\author{Andrew Du$^\dagger$ \quad Bo Chen$^\dagger$ \quad Tat-Jun Chin$^\dagger$ \quad Yee Wei Law$^\ddagger$ \quad Michele Sasdelli$^\dagger$ \\ 
\quad Ramesh Rajasegaran$^\dagger$ \quad  Dillon Campbell$^\dagger$ \\
$^\dagger$The University of Adelaide \quad $^\ddagger$University of South Australia}

\maketitle
%\thispagestyle{empty}

%%%%%%%%% ABSTRACT
\begin{abstract}
    Deep neural networks (DNNs) have become essential for processing the vast amounts of aerial imagery collected using earth-observing satellite platforms. However, DNNs are vulnerable towards adversarial examples, and it is expected that this weakness also plagues DNNs for aerial imagery. In this work, we demonstrate one of the first efforts on physical adversarial attacks on aerial imagery, whereby adversarial patches were optimised, fabricated and installed on or near target objects (cars) to significantly reduce the efficacy of an object detector applied on overhead images. Physical adversarial attacks on aerial images, particularly those captured from satellite platforms, are challenged by atmospheric factors (lighting, weather,  seasons) and the distance between the observer and target. To investigate the effects of these challenges, we devised novel experiments and metrics to evaluate the efficacy of physical adversarial attacks against object detectors in aerial scenes. Our results indicate the palpable threat posed by physical adversarial attacks towards DNNs for processing satellite imagery\footnote{See demo video at \url{https://youtu.be/5N6JDZf3pLQ}.}.
\end{abstract}

%%%%%%%%% BODY TEXT
%%%%%%%%%%%%%%%%%%%%%%%%%%%%%%%%%%%%%%%%%%%%%%%%%%%%%%%%%%%%%%%%%%%%%%%%%%%%%%%%%%%%%%%%%%
\section{Introduction}

The amount of images collected from earth-observing satellite platforms is growing rapidly~\cite{zhu2017deep}, in part fuelled by the dependency of many valuable applications on the availability of satellite imagery obtained in high cadence over large geographical areas, \eg, environmental monitoring~\cite{manfreda2018use}, urban planning~\cite{albert2017using}, economic forecasting~\cite{yeh2020using}. This naturally creates a need to automate the analysis of satellite or aerial imagery to cost effectively extract meaningful insights from the data. To this end, deep neural networks (DNNs) have proven effective~\cite{albert2017using,yeh2020using,zhu2017deep}, particularly for tasks such as image classification~\cite{albert2017using,pritt2017satellite}, object detection~\cite{ji2020vehicle,shermeyer2019effects}, and semantic segmentation~\cite{yuan2021review}.

However, the vulnerability of DNNs towards \emph{adversarial examples}, \ie, carefully crafted inputs aimed at fooling the models into making incorrect predictions, is well documented. Szegedy \etal \cite{szegedy2013intriguing} first showed that an input image perturbed with changes that are imperceptible to the human eye is capable of biasing convolutional neural networks (CNNs) to produce wrong labels with high confidence. Since then, numerous methods for generating adversarial examples~\cite{brendel2018decision, carlini2017towards, goodfellow2015explaining, kurakin2016adversarial, kurakin2017adversarial, madry2017towards, moosavi2017universal, moosavi2016deepfool, papernot2016transferability, papernot2017practical, papernot2016limitations, su2019one} and defending against adversarial attacks \cite{carlini2017magnet, goodfellow2015explaining, gu2014towards, papernot2016distillation, wang2016theoretical, xu2018feature} have been proposed. The important defence method of adversarial training \cite{goodfellow2015explaining, kannan2018adversarial, kurakin2017adversarial, madry2017towards} requires generating adversarial examples during training. Hence, establishing effective adversarial attacks is a crucial part of defence.

Adversarial attacks can be broadly categorised into \emph{digital attacks} and \emph{physical attacks}. Digital attacks directly manipulate the pixel values of the input images~\cite{akhtar2018threat, yuan2019adversarial}, which presume full access to the images. Hence the utility of digital attacks is mainly as generators of adversarial examples for adversarial training. Physical attacks insert real-world objects into the environment that, when imaged together with the targeted scene element, can bias DNN inference~\cite{athalye2017synthesizing, brown2017adversarial, eykholt2018robust, sharif2016accessorize, thys2019fooling}. The real-world objects are typically image patches whose patterns are optimised in the digital domain before being printed. A representative work is~\cite{thys2019fooling} who optimised patches to fool a person detector~\cite{redmon2016you}.

Our work focusses on physical adversarial attacks on aerial imagery, particularly earth-observing images acquired from satellite platforms. Previous works on physical attacks~\cite{athalye2017synthesizing, brown2017adversarial, eykholt2018robust, huang2020universal, komkov2021advhat, lee2019physical, sharif2016accessorize, thys2019fooling, wang2021dual, wang2021towards, wu2020making, xu2020adversarial, xue20203d}  overwhelmingly focussed on ground-based settings and applications, \eg, facial recognition, person detection and autonomous driving. A few exceptions include Czaja \etal~\cite{czaja2018adversarial}, who targeted aerial image classification, and den Hollander \etal~\cite{den2020adversarial}, who targeted aerial object detection. However, ~\cite{czaja2018adversarial,den2020adversarial} did not demonstrate their physical attacks in the real world, \ie, their patches were not printed and imaged from an aerial platform. Also, although~\cite{czaja2018adversarial} mentioned the potential effects of atmospheric factors on physical attacks, no significant investigation on this issue was reported.

\vspace{-1em} 

\begin{figure}
    \centering
    \begin{subfigure}[b]{0.80\columnwidth}
        \centering
        \includegraphics[width=\textwidth]{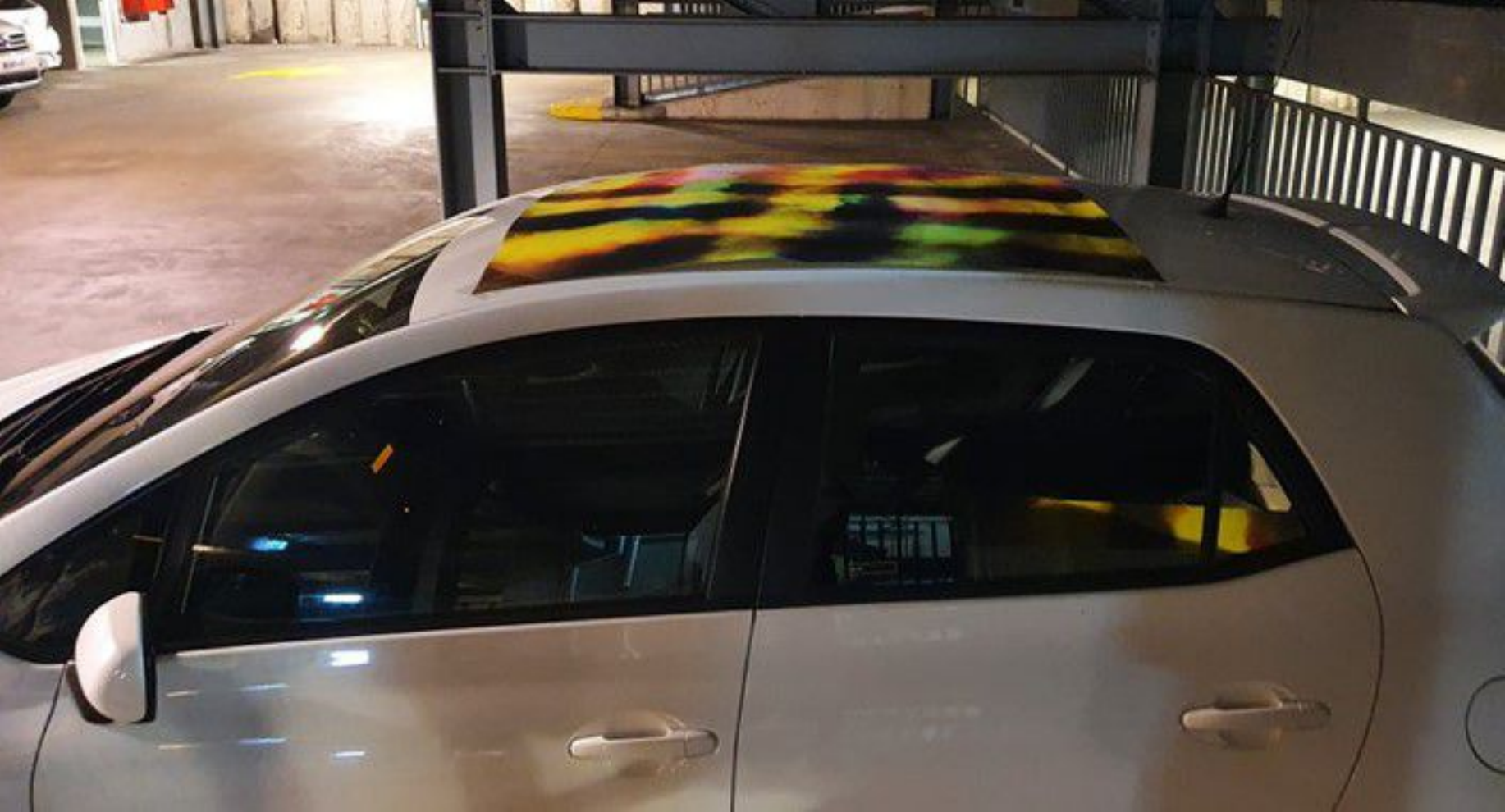}
        \caption{Adversarial patch on the roof of a car.}
        \label{fig:white_car}
    \end{subfigure}
    %\hfill
    \begin{subfigure}[b]{0.80\columnwidth}
        \centering
        \includegraphics[width=\textwidth]{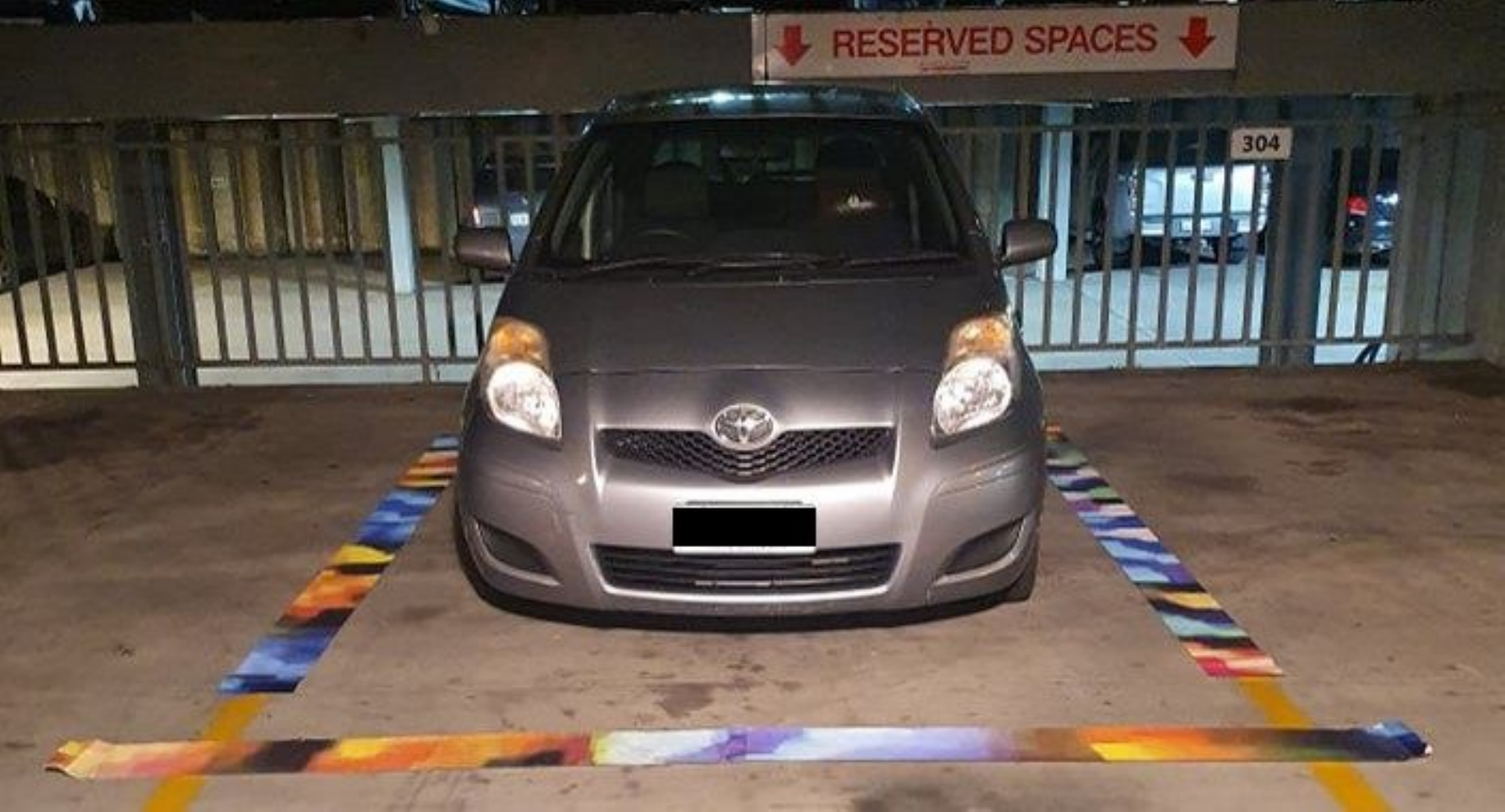}
        \caption{Adversarial patch off-and-around a car.}
        \label{fig:grey_car}
    \end{subfigure}
    \hfill
    \begin{subfigure}[t]{0.23\textwidth}
        \centering
        \includegraphics[width=\textwidth]{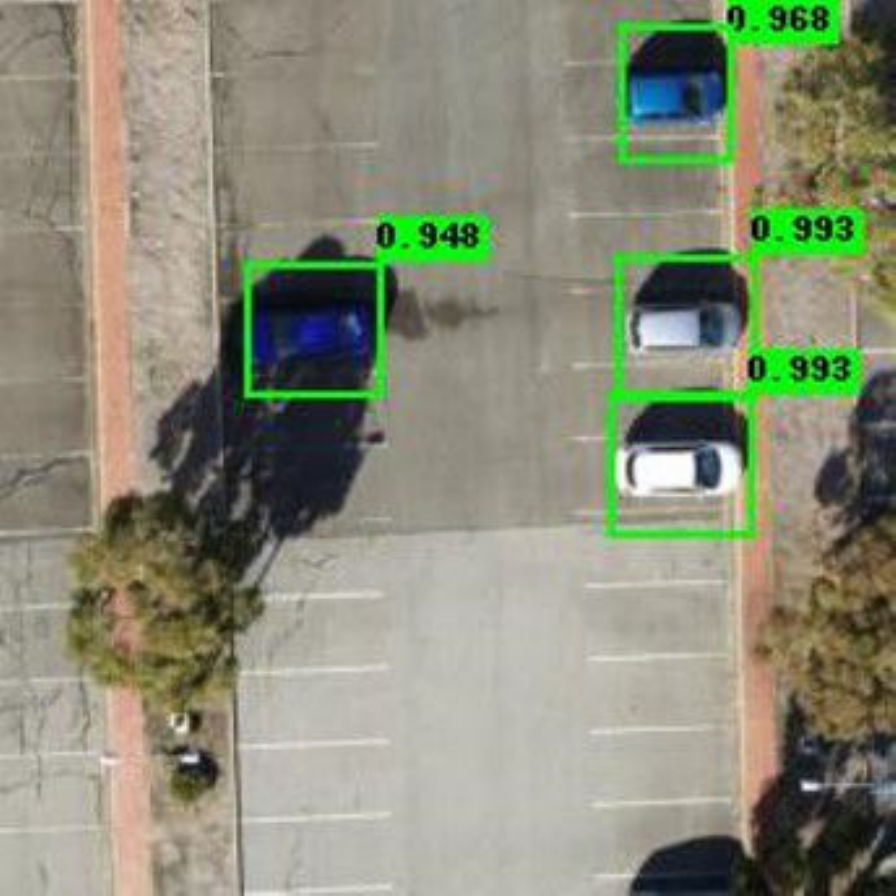}
        \caption{Cars in car park imaged without adversarial patches (objectness scores $\ge 0.9$).}
        \label{fig:carpark_nopatch}
    \end{subfigure}
    \hfill
    \begin{subfigure}[t]{0.23\textwidth}
        \centering
        \includegraphics[width=\textwidth]{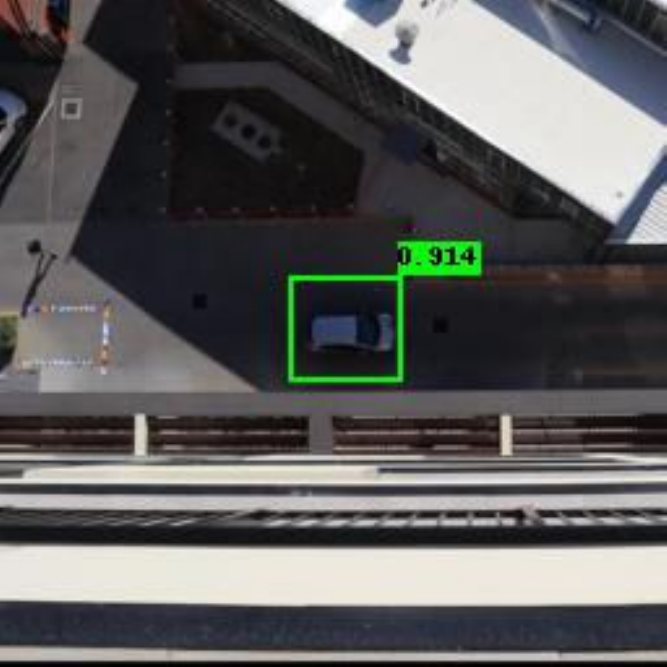}
        \caption{A car in side street imaged without adversarial patch (objectness score $ 0.914$).}
        \label{fig:sidestreet_nopatch}
    \end{subfigure}
    \hfill
    \begin{subfigure}[t]{0.23\textwidth}
        \centering
        \includegraphics[width=\textwidth]{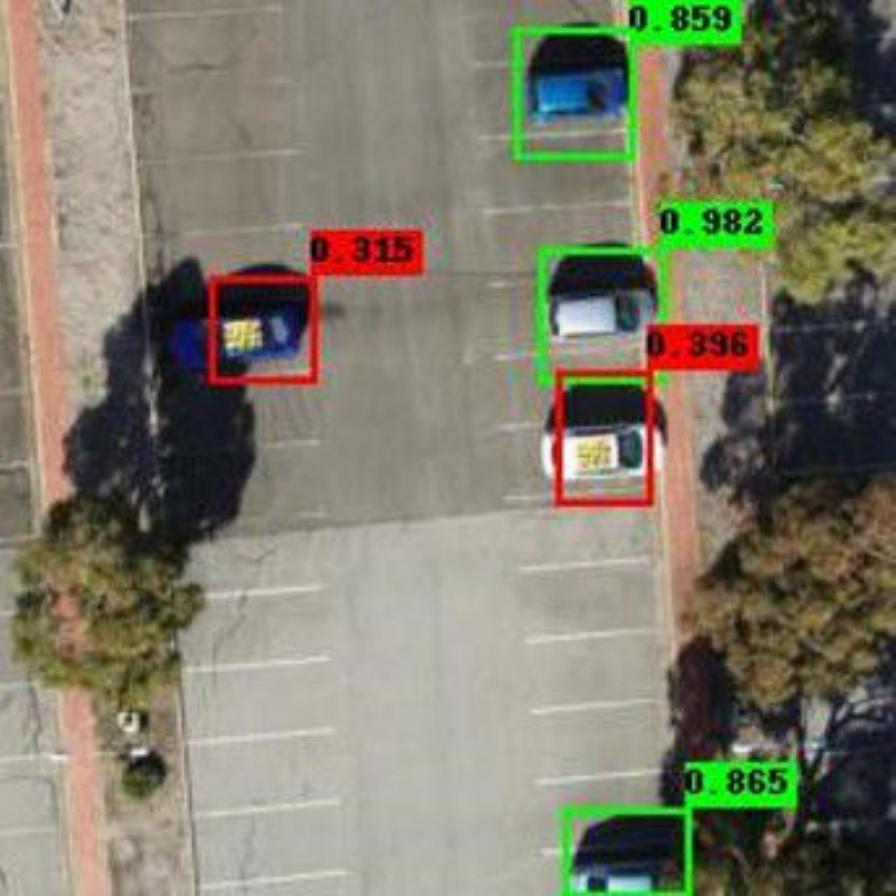}
        \caption{Two of the cars from (c) imaged with physical on-car patches; their objectness scores reduced to $0.315$ and $0.396$.}
        \label{fig:carpark_patch}
    \end{subfigure}     
    \hfill
    \begin{subfigure}[t]{0.23\textwidth}
        \centering
        \includegraphics[width=\textwidth]{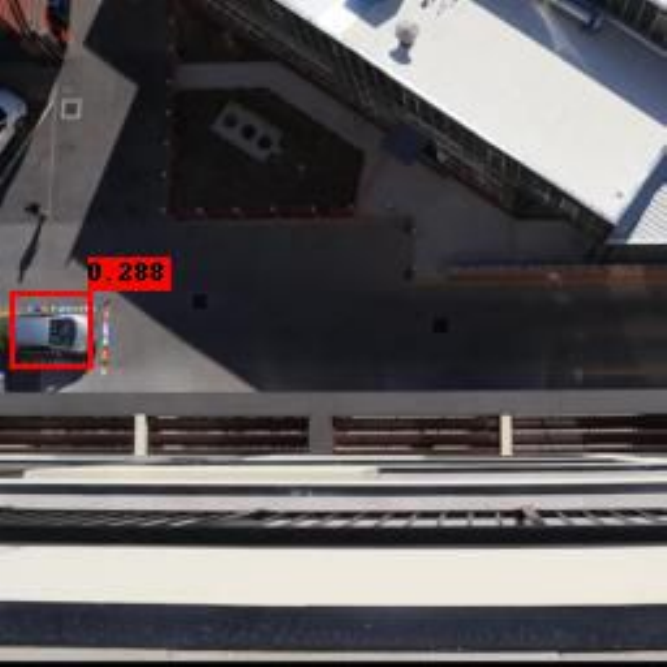}
        \caption{The car in (d) imaged with a physical off-car patch; the objectness score reduced to $0.288$.}
        \label{fig:sidestreet_offpatch}
    \end{subfigure}
    \vspace{-0.5em}
    \caption{Physical adversarial attacks on a YOLOv3 car detector in aerial imagery.}
    \label{fig:physical_attack}
\end{figure}

\paragraph{Contributions} In this applications paper, our contributions are threefold:

\begin{itemize}[leftmargin=1em,itemsep=2pt,parsep=0pt,topsep=2pt]
	\item We report one of the first demonstrations of real-world physical adversarial attacks in aerial scenes, specifically against a car detector. Adversarial patches were trained, printed and imaged from overhead using a UAV and from the balcony of a tall building; see Fig.~\ref{fig:physical_attack}. The captured images were processed using a car detector and the effectiveness of the physical attacks was then evaluated.
	\item We propose a novel adversarial patch design that surrounds the target object (car) with optimised intensity patterns (see Fig.~\ref{fig:grey_car}), which we also physically tested and evaluated. The design enables a more convenient attack since modifications to the car can be avoided, and the car can be driven directly into pattern on the ground.
	\item We examined the efficacy of physical attacks under different atmospheric factors (lighting, weather, seasons) that affect satellite imagery. To enable scalable evaluation, we designed an experimental protocol that performed data augmentation on aerial images and devised new, rigorous metrics to measure attack efficacy.
\end{itemize}

Our results indicate the realistic threat of physical attacks on aerial imagery object detectors. By making our code and data publicly available \cite{du2021physical}, we hope our work will spur more research into adversarial robustness of machine learning techniques for aerial imagery.
% \footnote{ \url{https://github.com/andrewpatrickdu/adv-yolov3-cowc}}

%%%%%%%%%%%%%%%%%%%%%%%%%%%%%%%%%%%%%%%%%%%%%%%%%%%%%%%%%%%%%%%%%%%%%%%%%%%%%%%%%%%%%%%%%
\section{Related work}\label{sec:related}

% Since our work is on physical adversarial attacks, our survey focusses on such attacks.
Three relevant areas of the literature on physical attacks are surveyed in this section.

\subsection{Physical attacks on image classification}

Kurakin \etal \cite{kurakin2016adversarial} generated the first physical-world attack by printing digitally perturbed images which were then captured by a smart phone and fed into an pre-trained Inception v3 \cite{szegedy2016rethinking} classifier. However, their results show that the effectiveness of the attack decreases when the images undergo the printing and photography processes. Lu \etal \cite{lu2017no} confirmed this loss of effectiveness when the images were viewed at different angles and distances.

Since then, a growing number of research papers have started proposing methods to generate adversarial examples that can survive in the physical world. For instance, Sharif \etal \cite{sharif2016accessorize} generated adversarial eyeglass frames to fool multiple facial recognition systems by adding a \emph{non-printability score} (NPS) and \emph{total variation} (TV) loss into their optimisation goal to ensure the colours used can be realised by a printer and the colour changes associated with adversarial perturbations are smooth. Komkov and Petiushko \cite{komkov2021advhat} added the TV loss into their optimisation goal to generate adversarial stickers (that were placed on hats) to fool a facial recognition system called ArcFace \cite{deng2019arcface}. 

Athalye \etal \cite{athalye2017synthesizing} proposed a method called \emph{Expectation Over Transformation} (EOT) which adds a transformation step into the optimisation process of generating adversarial perturbations. Their method was able to generate 3D-printed adversarial objects that were robust over a chosen distribution of synthetic transformations such as scale, rotation, translation, contrast, brightness, and random noise. 
% The EOT method has been shown to be very effective in a wide variety of attack scenarios. For instance, Brown \etal \cite{brown2017adversarial} generated adversarial patches which were placed next to the target object to fool various ImageNet \cite{deng2009imagenet} classifiers.
These patches were shown to be universal, \ie, can be used to attack any scene; robust, \ie, effective under a variety of transformations; and targeted, \ie, can cause a classifier to output any target class; as demonstrated in the ``rogue'' traffic sign scenarios \cite{sitawarin2018darts,sitawarin2018rogue}, and even scenarios where the patch is adjacent to rather than on the targeted object \cite{brown2017adversarial}.
% Sitawarin \etal \cite{sitawarin2018darts, sitawarin2018rogue} used the EOT method to generate adversarial traffic signs. 
Eykholt \etal \cite{eykholt2018robust} proposed a refinement of the EOT method called \emph{Robust Physical Perturbation} (RP2), which performs sampling over a distribution of synthetic and physical transformations to generate adversarial stop signs in the form of posters or black-and-white stickers to be placed on stop signs. However, in order to use the RP2 method, the attacker would need to print out the original (clean) image and then, take multiple photos of it from different angles and distances to generate a single adversarial example. Jan \etal \cite{jan2019connecting} proposed a transformation called D2P, to be performed prior to EOT, which uses a cGAN \cite{isola2017image, zhu2017unpaired} to simulate the effects produced by the printing and photography process. Their method also suffers from a feasibility problem since the attacker would need to print out hundreds of images and then capture them with a camera to build the ground truth to train the network.   

\subsection{Physical attacks on object detection}

The building blocks of adversarial patch generation discussed in the previous subsection, ranging from NPS minimisation to cGAN, are also applicable to physical attacks on object detection, \eg, detection of traffic signs \cite{eykholt2017note, lu2017adversarial, chen2018shapeshifter, song2018physical}, which is of practical importance considering the emergence of autonomous driving.
% Lu \etal \cite{lu2017adversarial}, Eykholt \etal \cite{eykholt2017note}, Chen \etal \cite{chen2018shapeshifter} and Song \etal \cite{song2018physical} for example used some of the aforementioned building blocks to generate adversarial stop signs to fool the YOLOv2/YOLO9000 \cite{redmon2017yolo9000} and Faster R-CNN \cite{ren2015faster} object detectors.
These early work targeted YOLOv2/YOLO9000 \cite{redmon2017yolo9000} and Faster R-CNN \cite{ren2015faster} object detectors.
% Xue \etal \cite{xue2021naturalae} also generated adversarial stop signs and demonstrated the transferability of their attack across a variety of object detectors.
Adversarial examples trained using YOLOv2 do not transfer well to (i.e., is not as effective against) Faster R-CNN, and vice versa, but adversarial examples trained using either YOLOv2 or Raster R-CNN transfer well to single-shot detectors \cite{xue2021naturalae}.

Thys \etal \cite{thys2019fooling} generated adversarial cardboard patches to hide people from a YOLOv2-based detector, but attacks targeting YOLOv2 do not transfer well to YOLOv3 \cite{redmon2018yolov3} because YOLOv3 supports multi-label prediction and can detect small objects better~\cite{wang2021towards}.
% = = = = Andrew's original = = = =
% Wang \etal \cite{wang2021towards} extended this work by attacking a YOLOv3 \cite{redmon2018yolov3} and Faster R-CNN \cite{ren2015faster} detector as well as demonstrating the transferability of the attack across a variety of object detectors trained on different datasets.
% Lee and Kolter \cite{lee2019physical} showed that adversarial patches are also effective when placed away from the target object. Xu \etal \cite{xu2020adversarial}, Wu \etal \cite{wu2020making}, Huang \etal \cite{huang2020universal}, and Xue \etal \cite{xue20203d} printed adversarial patches onto clothing to either hide a person or cause a detector to misclassify them as some target class.
% = = = = Law's notes = = = =
% \cite{wu2020making} uses ensemble training to simultaneously target multiple object detectors, e.g., YOLOv2 + Faster R-CNN with a ResNet-50 backbone and multi-scale features. Effective against either of these but not YOLOv3 for example.
% \cite{lee2019physical} experimented with off-body patches to hide people from YOLOv3.
% \cite{xu2020adversarial} takes non-rigid deformation into account through thin plate spline mapping. YOLOv2.
% \cite{huang2020universal} trains patches on large surface areas against variants of Faster R-CNN. High viewing angle and large distance make attacks less effective --- same problem \cite{xue20203d} has.
% \cite{xue20203d} still targets YOLOv2 but addresses 3D-like transformations in terms of viewing angle, wrinkling, stereoscopic radians, occlusion.
From \cite{thys2019fooling}, several directions emerged: (i) off-body patches \cite{lee2019physical}; (ii) physical adversarial examples that are robust to physical-world transformations, taking into account non-rigid deformations of patches \cite{xu2020adversarial}, viewing angle, wrinkling, stereoscopic radians, occlusion \cite{xue20203d}, material constraint, semantic constraint \cite{huang2020universal}, \etc; (iii) targeting more advanced detectors than YOLOv2 such as YOLOv3 \cite{lee2019physical, wu2020making}, and multiple detectors simultaneously using ensemble training \cite{wu2020making}.

% Interestingly, Zhang \etal \cite{zhang2019camou} and Wang \etal \cite{wang2021dual} generated adversarial patches (in the form of camouflage) to hide 3D model cars from being detected in a simulation environment based on the Unreal Engine.
Cars have been the targeted objects recently \cite{zhang2019camou, wang2021dual}, but in these prior work, simulators based on the Unreal Engine and at most toy cars were used. Moreover, except for unpaintable surfaces like windows, the entire car was subjected to adversarial patching, whereas in our case, adversarial patches are subject to size constraints.

\subsection{Physical attacks in aerial imagery}

Physical attacks have mostly been demonstrated on earth-based imagery captured at relatively small distances \eg, within the sensing range of a camera on an autonomous car or a security system. Physical attacks against aerial or satellite imagery have not been considered or extensively studied. A few exceptions are Czaja \etal~\cite{czaja2018adversarial} and den Hollander \etal~\cite{den2020adversarial}, who generated adversarial patches for aerial imagery classification and detection respectively. However,~\cite{czaja2018adversarial,den2020adversarial} only evaluated their attacks digitally, \ie, they did not print their patches and deploy them in the real world. Robustifying physical attacks againts atmospheric effects, temporal variability and material properties was mentioned but no significant work was reported in~\cite{czaja2018adversarial}.

%%%%%%%%%%%%%%%%%%%%%%%%%%%%%%%%%%%%%%%%%%%%%%%%%%%%%%%%%%%%%%%%%%%%%%%%%%%%%%%%%%%%%%%%%%
\section{Threat model}\label{sec:threat_model} 

We first present the threat model used in our work.

\textbf{Attacker's goals:} The attacker aims to generate adversarial patches that can prevent cars existing in a selected locality, called the \textbf{scene of attack}, from being detected from the air by a pre-trained car detector. These patches are to be placed on the roof of cars, or off-and-around cars; see Figs.~\ref{fig:white_car} and \ref{fig:grey_car}. Although the attacker is only interested in hiding cars in the scene of attack, the patches should still be effective in different environmental conditions.

\textbf{Attacker's knowledge:} The attacker is assumed to have \emph{white-box} (\ie, complete) knowledge of the detector model
% such as training data, model architecture, training hyperparameters, loss function minimised during training, trained model parameters (weights and biases), \etc.
including its parameter values, architecture, loss function, optimiser, and in some cases its training data as well \cite[p. 22]{nistir8269}.
% This is a reasonable assumption to make since physical adversarial examples generated on one detector have been shown to transfer well to other detectors \cite{lu2017adversarial, song2018physical, wu2020making}.
Examples of scenarios where this assumption is valid include when (i) the targeted detector is known to be taken from some open-source implementation, (ii) the attacker can access and reverse-engineer a black-box detector implementation. More importantly, this assumption represents the worst-case scenario for the defender, which allows us to assess the maximum impact the attacker can cause. 
% (i) the attacker is an insider, 
% The transferability of the white-box attacks to be discussed is of interest, but is not a focus of the current work. Furthermore, ensemble training is known to enable an attack to target multiple detectors \cite{wu2020making}.

\textbf{Attacker's capabilities:} The attacker can optimise and evaluate the adversarial patches in the digital domain, and perform physical-world attacks by printing the patches and placing them physically in the scene of attack.  

\textbf{Attacker's strategy:} The attacker will attempt to generate adversarial patches by solving an optimisation problem that includes the maximum objectness score in the loss function. See Sec.~\ref{sec:patch} for more details. 

%%%%%%%%%%%%%%%%%%%%%%%%%%%%%%%%%%%%%%%%%%%%%%%%%%%%%%%%%%%%%%%%%%%%%%%%%%%%%%%%%%%%%%%%%%
\section{Methods}\label{sec:methods} 

Based on the threat model in Sec.~\ref{sec:threat_model}, this section discusses the optimisation of physical adversarial patches to attack car detectors in aerial imagery.

\subsection{Object detector model}\label{sec:yolov3}

\begin{figure}
     \centering
     \begin{subfigure}[t]{0.23\textwidth}
         \centering
         \includegraphics[width=\textwidth]{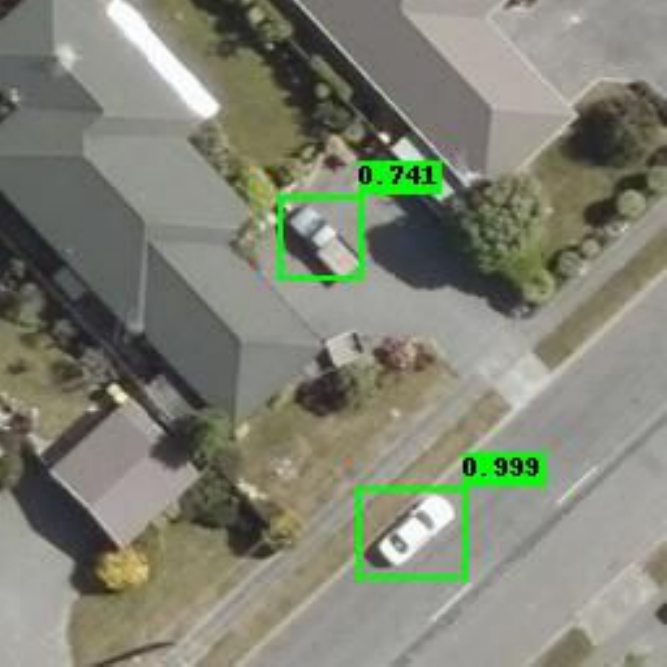}
         \caption{COWC - Selwyn}
         \label{fig:cowc1}
     \end{subfigure}
     \hfill
     \begin{subfigure}[t]{0.23\textwidth}
         \centering
         \includegraphics[width=\textwidth]{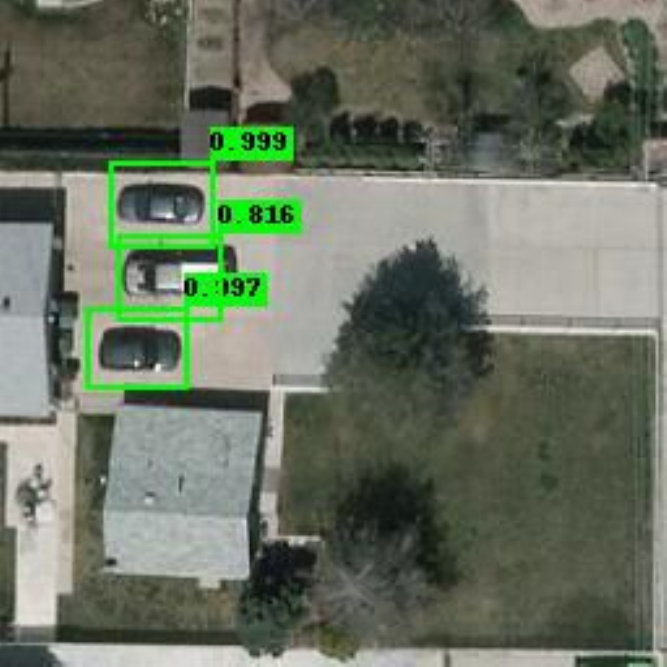}
         \caption{COWC - Utah}
         \label{fig:cowc2}
     \end{subfigure}
     \hfill
     \begin{subfigure}[t]{0.23\textwidth}
         \centering
         \includegraphics[width=\textwidth]{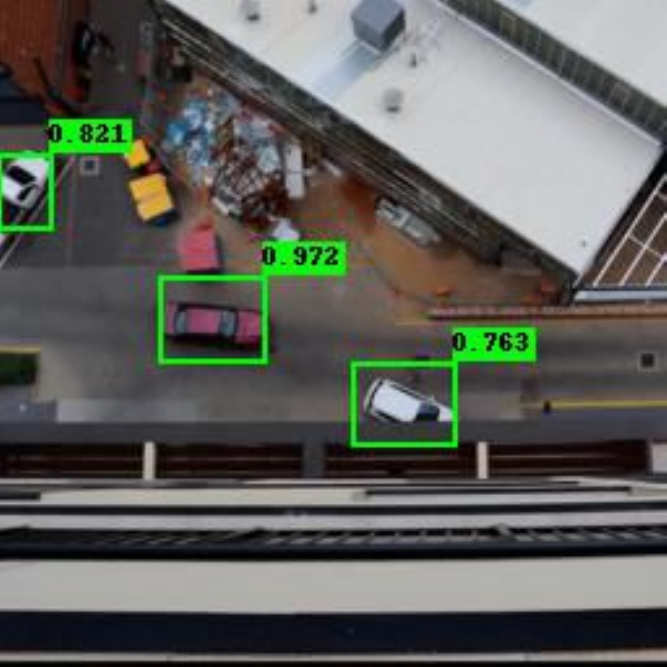}
         \caption{Side Street}
         \label{fig:sidestreet1}
     \end{subfigure}
     \hfill
     \begin{subfigure}[t]{0.23\textwidth}
         \centering
         \includegraphics[width=\textwidth]{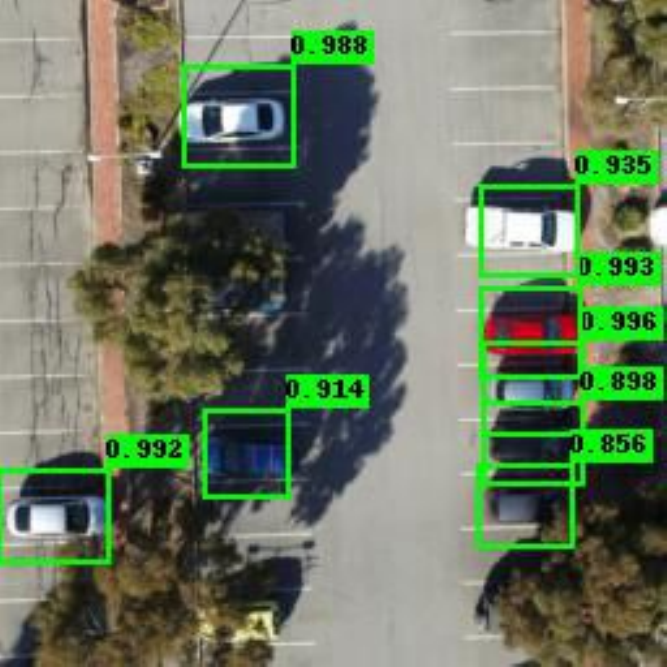}
         \caption{Car Park}
         \label{fig:carpark1}
     \end{subfigure}
        \caption{Overhead view of cars from three datasets.}
        \label{fig:dataset}
\end{figure}

To build a car detector for aerial imagery, we performed transfer learning on YOLOv3~\cite{redmon2018yolov3} (pretrained on MS-COCO~\cite{lin2014microsoft}) on the Cars Overhead with Context (COWC) dataset~\cite{mundhenk2016large}. We used a version of COWC called COWC-M which contained $25,384$ colour images ($256 \times 256$ pixels) of annotated cars in an overhead view (Figs.~\ref{fig:cowc1} and \ref{fig:cowc2}). We separated the data into $20,306$ training and $5,078$ testing samples. The mean Average Precision (mAP) measured at IOU threshold of $0.50$ of our detector was $0.50$, which was on par with YOLOv3 trained on MS-COCO.

\subsection{Scenes of attack and data collection}\label{sec:data-collection}

Following the threat model in Sec.~\ref{sec:threat_model}, we selected and collected data from two scenes of attack:
\begin{description}[leftmargin=1em,itemsep=2pt,parsep=0pt,topsep=2pt]
\item[Side Street] Street viewed from the 10th floor (40 m height) of a building (see Fig.~\ref{fig:sidestreet1}). Data for training adversarial patches were captured with a Canon EOS M50 camera with a 15-45 mm lens. A total of $843$ images were captured and divided into $780$ training and $63$ testing images. 

\item[Car Park] Parking area with approximately 400 parking spaces (see Fig.~\ref{fig:carpark1}). Data for training adversarial patches were captured with a DJI Zenmuse X7 camera with a 24 mm lens on a DJI Matrice 200 Series V2 UAV at a fixed height of 60 m. A total of $565$ images were captured and separated into $500$ training and $65$ testing images. 
\end{description}
The data collected from the selected scenes sufficiently resembles the COWC data and the pre-trained car detector worked successfully on the collected images; see Fig.~\ref{fig:dataset}.

\vspace{-1em}

\paragraph{Annotation for patch optimisation} Let $\cU = \{ I_i \}^{M}_{i=1}$ and $\cV = \{ J_j \}^{N}_{j=1}$ respectively be the $M$ training and $N$ testing images for a particular scene of attack. We executed the YOLOv3 car detector (Sec.~\ref{sec:yolov3}) on $\cU$ and $\cV$ with an objectness threshold of $0.5$ and non-max suppression IoU threshold of $0.4$. This yields the tuples
\begin{align}
    \mathcal{T}_i^{\cU} = \{ (\mathcal{B}_{i,k}^{\cU}, s_{i,k}^{\cU}) \}^{D_i}_{k=1}, \;\; \mathcal{T}_j^{\cV} = \{ (\mathcal{B}_{j,\ell}^{\cV}, s_{j,\ell}^{\cV}) \}^{E_j}_{\ell=1}
\end{align}
for each $I_i$ and $J_j$, where
\begin{itemize}[leftmargin=1em,itemsep=2pt,parsep=0pt,topsep=2pt]
\item $D_i$ is the number of detections in $I_i$;
\item $\mathcal{B}_{i,k}^{\cU}$ is the bounding box of the $k$-th detection in $I_i$;
\item $s_{i,k}^{\cU}$ is the objectness score of the $k$-th detection in $I_i$
\end{itemize}
(similarly for $E_j$, $\mathcal{B}_{j,\ell}^{\cV}$ and $s_{j,\ell}^{\cV}$ for $J_j$). Note that by design $s_{i,k}^{\cU} \ge 0.5$ and $s_{j,\ell}^{\cV} \ge 0.5$. The sets of all detections are
\begin{align}
\mathcal{T}^{\cU} = \{ \mathcal{T}^{\cU}_i \}^{M}_{i=1}, \;\; \mathcal{T}^{\cV} = \{ \mathcal{T}^{\cV}_j \}^{N}_{j=1},
\end{align}
which we manually checked to remove false positives.

\subsection{Optimising adversarial patches}\label{sec:patch}

\begin{figure}[h]
    \centering
    \begin{subfigure}[b]{0.20\textwidth}
        \centering
        \includegraphics[width=\textwidth]{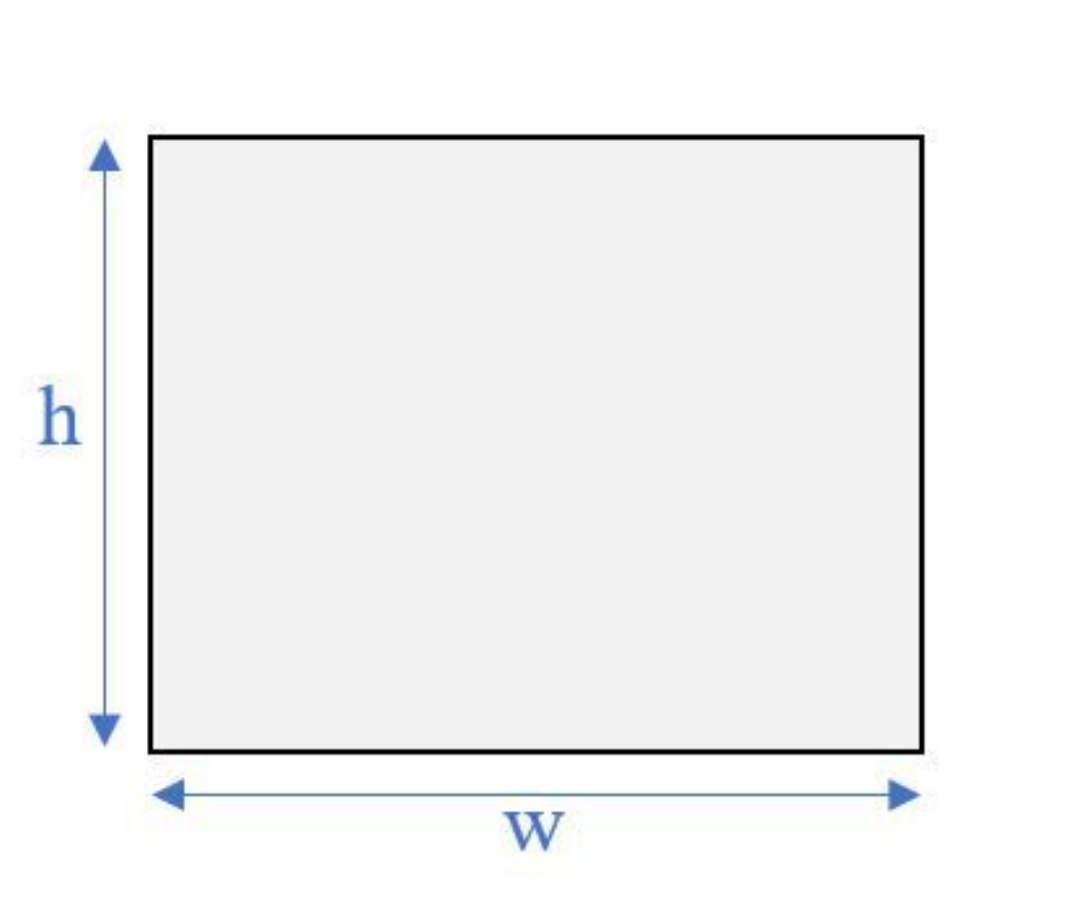}
        \caption{Type ON.}
    \end{subfigure}
    \hspace{1em}
    \begin{subfigure}[b]{0.20\textwidth}
        \centering
        \includegraphics[width=\textwidth]{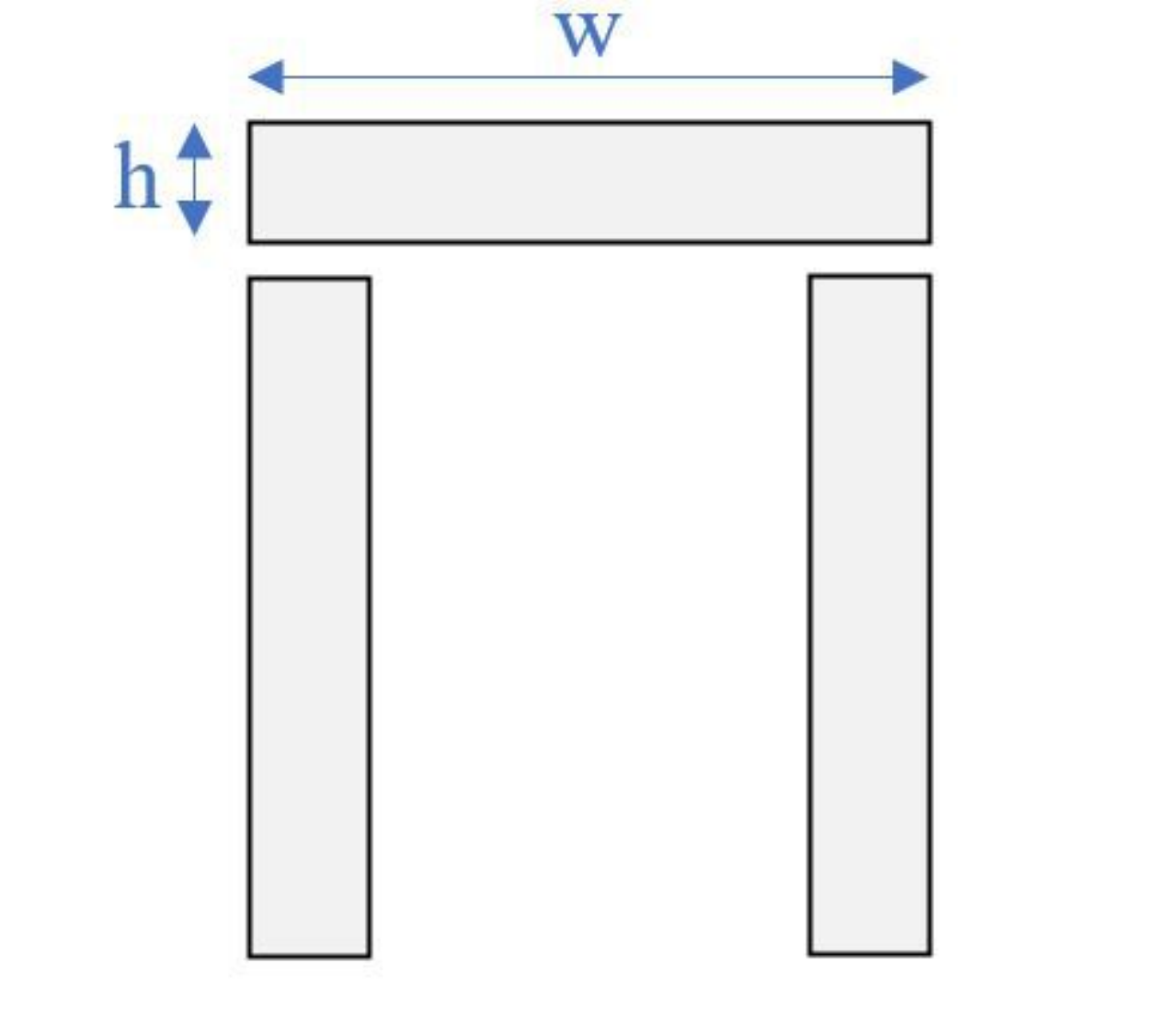}
        \caption{Type OFF.}
    \end{subfigure}
    \vspace{-1em}
    \caption{Patch designs.}
    \label{fig:designs}
\end{figure}

\begin{figure*}[ht]\centering
	 \includegraphics[width=1.0\linewidth]{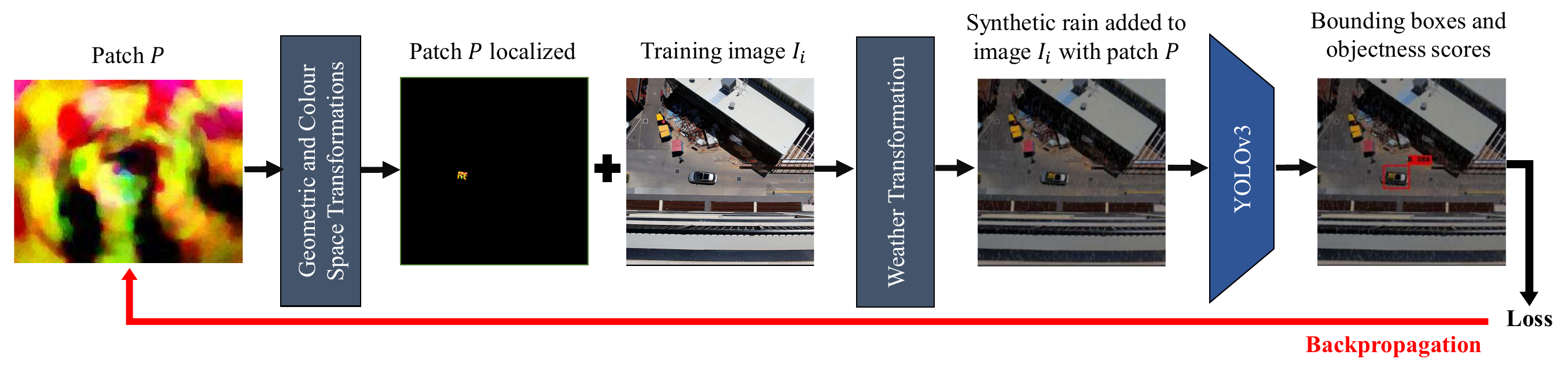}
	 \vspace{-2em}
	 \caption{Optimisation process for generating adversarial patches.}
	 \label{fig:pipeline}
\end{figure*}

We adapted Thys \etal's~\cite{thys2019fooling} method for adversarial patch optimisation for aerial scenes; see Fig.~\ref{fig:pipeline} for our pipeline. Two patch designs were considered (see Fig.~\ref{fig:designs}):
\begin{itemize}[leftmargin=1em,itemsep=2pt,parsep=0pt,topsep=2pt]
    \item Type ON: rectangular patch to be installed on car roof. The following dimensions were used:
\begin{itemize}[leftmargin=1em,itemsep=2pt,parsep=0pt,topsep=2pt]
\item Digital dimensions: $w = 200$ pixels, $h = 160$ pixels
%\item Digital dims (Car park): $w = 300$ pix, $h = 240$ pix
\item Physical dimensions: $w = 1189$ mm, $h = 841$ mm
\end{itemize}    
    \item Type OFF: three rectangular strips to be installed off and around car, forming a `$\sqcap$' shape, with dimensions:
\begin{itemize}[leftmargin=1em,itemsep=2pt,parsep=0pt,topsep=2pt]
\item Digital dimensions: $w = 400$ pixels, $h = 25$ pixels
%\item Digital dims (Car park): 
\item Physical dimensions: $w = 3200$ mm, $h = 200$ mm
\end{itemize}    
\end{itemize}
See Fig.~\ref{fig:physical_attack} for the physical placement of the patches.

\vspace{-1em}

Let $P$ be the set of colour pixels that define a digital patch (ON or OFF). The data used to optimise $P$ for a scene of attack are the training images $\cU$ and annotations $\mathcal{T}^{\mathcal{U}}$ for the scene. The values of $P$ are initialised randomly. Given the current $P$, the patch is embedded into each training image $I_i$ based on the bounding boxes $\{ \cB^\cU_{i,k} \}^{D_i}_{k=1}$.

Several geometric and colour-space augmentations are applied to $P$ to simulate its appearance when captured in the field. The geometric transformations applied are:
\begin{itemize}[leftmargin=1em,itemsep=2pt,parsep=0pt,topsep=2pt]
    \item Random scaling of the embedded $P$ to a size that is roughly equivalent to the physical size of $P$ in the scene.
    \item Random rotations ($\pm 20^\circ$) on the embedded $P$ about the centre of the bounding boxes $\{ \cB^\cU_{i,k} \}^{D_i}_{k=1}$.
\end{itemize}
The above simulate placement and printing size uncertainties. The colour space transformations are
\begin{itemize}[leftmargin=1em,itemsep=2pt,parsep=0pt,topsep=2pt]
    \item Random brightness adjustment ($\pm 0.1$ change of pixel intensity values).
    \item Random contrast adjustment ($[0.8, 1.2]$ change of pixel intensity values).
    \item Random noise ($\pm 0.1$ change of pixel intensity values).
\end{itemize}
The above simulate (lack of) colour fidelity of the printing device and lighting conditions in the field.

\begin{figure}[ht]\centering
     \begin{subfigure}[b]{0.23\textwidth}
         \centering
         \includegraphics[width=\textwidth]{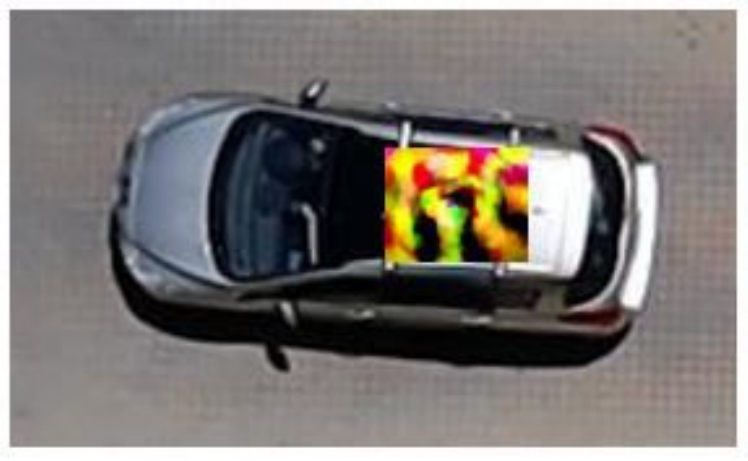}
         \caption{No geometric and colour-space transformations.}
         \label{fig:gc_off}
     \end{subfigure}
     \hfill
     \begin{subfigure}[b]{0.23\textwidth}
         \centering
         \includegraphics[width=\textwidth]{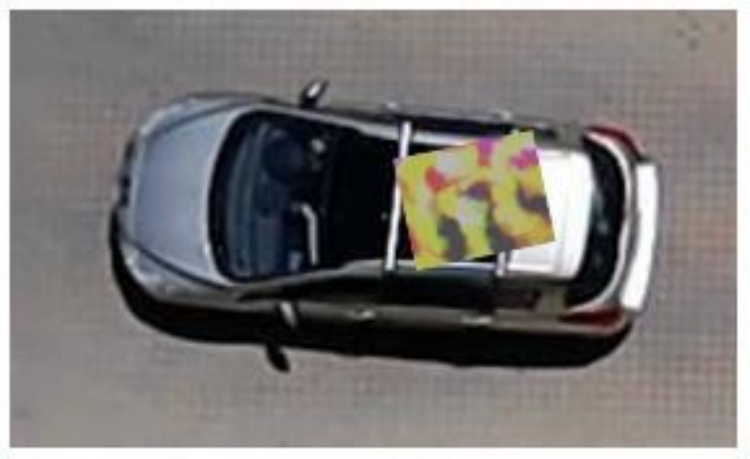}
         \caption{With geometric and colour-space transformations.}
         \label{fig:gc_on}
     \end{subfigure}
     \hfill
     \begin{subfigure}[b]{0.23\textwidth}
         \centering
         \includegraphics[width=\textwidth]{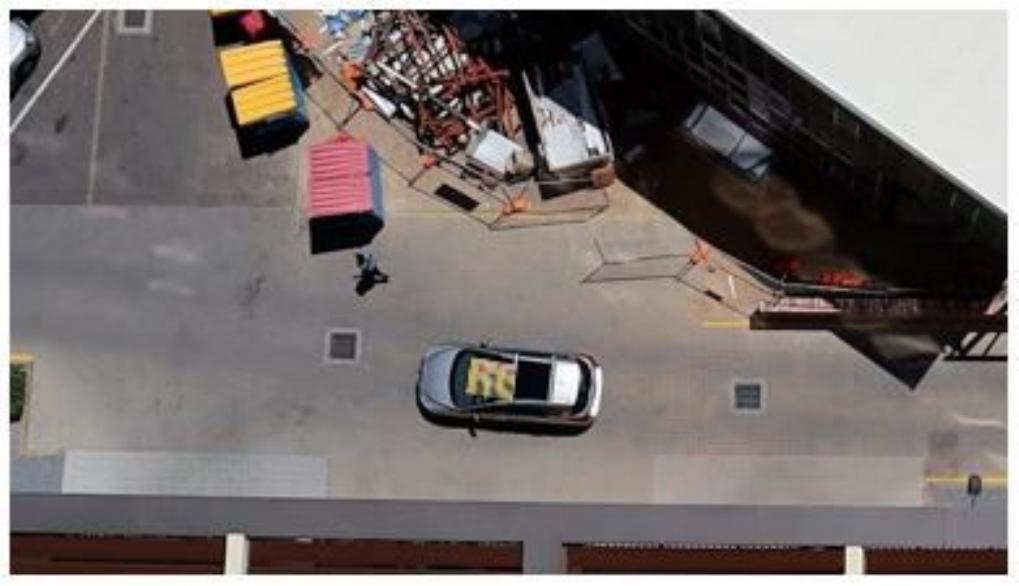}
         \caption{No weather transformation.}
         \label{fig:no_weather_transform}
     \end{subfigure}
     \hfill
     \begin{subfigure}[b]{0.23\textwidth}
         \centering
         \includegraphics[width=\textwidth]{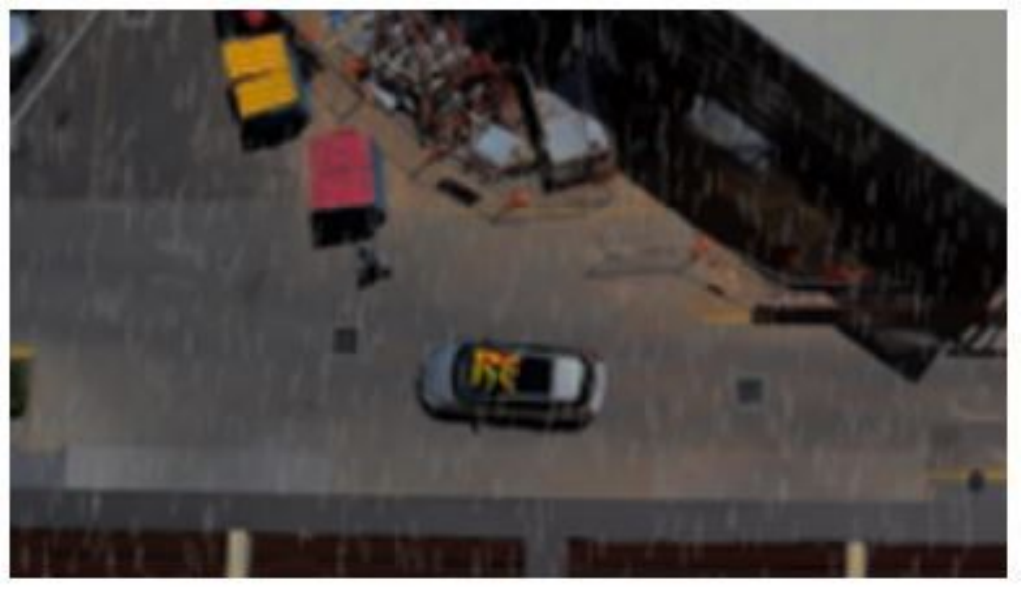}
         \caption{Synthetic rain added.}
         \label{fig:synthetic_rain_add}
     \end{subfigure}
        \caption{Effects of augmentations applied in our pipeline.}
        \label{fig:augmentations}
\end{figure}

Let $\tilde{I}_i$ be $I_i$ embedded with $P$ following the steps above. Since the patches are situated outdoors, we also considered weather and seasonal changes. The Automold tool~\cite{automold2018} was applied on each $\tilde{I}_i$ to adjust the sun brightness and add weather and seasonal effects (snow, rain, fog, autumn leaves). See Fig.~\ref{fig:augmentations} for the augmentations applied. Ablation studies on the augmentations will be presented in Sec.~\ref{sec:results}.

The augmented image $\tilde{I}_i$ is forward propagated through the car detector (Sec.~\ref{sec:yolov3}). The output consists of three grids which correspond to the three prediction scales of YOLOv3. Each cell in the output grids contains three bounding box predictions with objectness scores. Let $\tilde{\cS}_i^{\cU}$ be the set of all predicted objectness scores for $\tilde{I}_i$. Since we aim to prevent cars from being detected, we define the loss for $\tilde{I}_i$ as
\begin{equation}
    L_i(P) = \max \left( \tilde{\cS}_i^{\cU} \right) + \delta \cdot NPS(P) + \gamma \cdot TV(P), \label{eq:loss}
\end{equation}
where $\max ( \tilde{\cS}_i^{\cU} ) $ is the maximum predicted objectness score in $\tilde{I}_i$, and $\delta$, $\gamma$ are weights for the non-printability score~\cite{sharif2016accessorize} and total variation~\cite{sharif2016accessorize}  of $P$ respectively. Specifically,
\begin{align}
    NPS(P) = \sum_{u,v} \left( \min_{c \in C} \left\| \mathbf{p}_{u,v} - \mathbf{c}\right\|_2 \right),
\end{align}
where $\mathbf{p}_{u,v}$ is the pixel (RGB vector) at $(u,v)$ in $P$, and $\mathbf{c}$ is a colour vector from the set of printable colours~\cite{thys2019fooling}. 
% Removed symbol $C$ since it's used only once, and a reviewer asked about it.
The NPS term encourages colours in $P$ to be as close as possible to colours that can be reproduced by a printing device. The TV term
% \begin{align*}\label{eq:tv}
%     TV(P) = \sum_{u,v} \sqrt{(P_{u,v} - P_{u+1,v})^2 + (P_{u,v} - P_{u,v+1})^2}
% \end{align*}
% encourages $P$'s that are smooth, where $P_{u,v}$ is the intensity of pixel $(u,v)$, which also contributes to the physical realisability of $P$~\cite{thys2019fooling}. Minimising $L$ to optimise $P$ is achieved using the Adam~\cite{kingma2014adam} stochastic optimisation algorithm (note the pre-trained detector model is not updated).
\begin{align*}\label{eq:tv}
    TV(P) = \sum_{t,u,v} \sqrt{(p_{t,u,v} - p_{t,u+1,v})^2 + (p_{t,u,v} - p_{t,u,v+1})^2}
\end{align*}
encourages $P$'s that are smooth, where $p_{t,u,v}$ is the $t$-channel pixel (scalar) at $(u,v)$ in $P$, which also contributes to the physical realisability of $P$. Minimising $L$ to optimise $P$ is achieved using the Adam~\cite{kingma2014adam} stochastic optimisation algorithm. Note that the pre-trained detector is not updated.

%%%%%%%%%%%%%%%%%%%%%%%%%%%%%%%%%%%%%%%%%%%%%%%%%%%%%%%%%%%%%%%%%%%%%%%%%%%%%%%%%%%%%%%%%
\section{Measuring efficacy of physical attacks}\label{sec:metrics}

The efficacy of the patch $P^\ast$ optimised according to Sec.~\ref{sec:methods} is to be evaluated, but previous metrics are not directly or intuitively interpretable here. For example, \emph{average precision}~\cite{den2020adversarial, thys2019fooling, wu2020making} measures the classification accuracy of object detectors, whereas \emph{attack success rate}~\cite{czaja2018adversarial, eykholt2018robust, wu2020making, xu2020adversarial} focusses on misclassification. While these metrics can be applied in our experiments\footnote{See supplementary material in Sec.~\ref{sec:supp} for our average precision results.}, we propose new metrics that directly measure the impact on objectness score, developed according to two evaluation regimes.
% See supplementary material for alternative versions of our results expressed in precision-recall curves and average precision values.
% and \emph{fooling rate}~\cite{wang2021towards}

\subsection{Evaluation in the digital domain}\label{sec:aorr}

As a sanity test, as well as a more scalable approach to perform ablation studies, we evaluate $P^\ast$ using the testing set $\cV$ from Sec.~\ref{sec:data-collection}. In addition to the detection results $\cT^\cV$ on $\cV$, we embed $P^\ast$ into each $J_j \in \cV$ (following the same augmentation steps in Sec.~\ref{sec:patch}) to yield $\tilde{J}_j$. Executing the car detector (Sec.~\ref{sec:yolov3}) on $\tilde{J}_j$ yields the tuple
\begin{align}
    \tilde{\mathcal{T}}_j^{\cV} = \{ (\tilde{\mathcal{B}}_{j,\ell}^{\cV}, \tilde{s}_{j,\ell}^{\cV}) \}^{E_j}_{\ell=1}.
\end{align}
A one-to-one matching between $\mathcal{T}_j^{\cV}$ and $\tilde{\mathcal{T}}_j^{\cV}$ is achieved by using an objectness threshold close to zero, and associating for each $\mathcal{B}_{j,\ell}^{\cV}$ the resulting bounding box with the highest objectness score that overlaps sufficiently with $\mathcal{B}_{j,\ell}^{\cV}$. Let
\begin{align}
    \tilde{\mathcal{T}}^{\cV} = \{ \tilde{\mathcal{T}}_j^{\cV} \}^{N}_{j=1}.
\end{align}
Define \emph{average objectness reduction rate (AORR)} as
\begin{equation}\label{eq:AORR}
    \text{AORR}(\mathcal{T}^{\cV},\tilde{\mathcal{T}}^{\cV}) \triangleq
        \frac{1}{N E_j} \left( \sum^{N}_{j=1}  \sum^{E_j}_{\ell=1}   
        \frac{s_{j,\ell}^{\cV} - \tilde{s}_{j,\ell}^{\cV}}
             {s_{j,\ell}^{\cV}}\right).
\end{equation}
Intuitively, AORR measures the average reduction in objectness score of an object due to the adversarial patch. A more effective attack will yield a higher AORR.

\subsection{Evaluation in the physical domain}\label{sec:osr_ndr}

To evaluate $P^\ast$ in the physical domain, we collect new image sets (e.g., videos) $\cF = \{ F_f \}^{\alpha}_{f=1}$ and $\cG = \{ G_g \}^{\beta}_{g=1}$ from the same scene of attack, with $\cF$ containing the physical realisations of $P^\ast$ installed on several target cars, while $\cG$ does not contain any adversarial patches.

For each car that was targeted in $\cF$, the car was tracked using the Computer Vision Annotation Tool (CVAT)~\cite{cvat} in both $\cF$ and $\cG$. Both sets $\cF$ and $\cG$ were then subject to the car detector (Sec.~\ref{sec:yolov3}), and the objectness scores corresponding to the targeted object
\begin{align}
\tilde{\cS} = \{ \tilde{s}_f \}^{\alpha}_{f=1}, \;\; \cS = \{ s_g \}^{\beta}_{g=1}
\end{align}
were retrieved (similarly to Sec.~\ref{sec:aorr}). Define the \emph{objectness score ratio (OSR)} of the targeted car as
\begin{equation}\label{eq:OSR}
    \text{OSR}(\tilde{\cS},\cS) \triangleq
        \frac{\frac{1}{\alpha} \sum^{\alpha}_{f=1} \tilde{s}_f }
             {\frac{1}{\beta} \sum^{\beta}_{g=1} s_g }.
\end{equation}
Intuitively, OSR measures the ratio of the objectness scores after and before the application of physical patch $P^\ast$. A more effective attack will yield a lower OSR.

Define the \emph{normalised detection rate (NDR)} of the targeted car as

\begin{equation}\label{eq:NDR}
    \text{NDR}_\tau(\tilde{\cS},\cS) \triangleq 
    \frac{\frac{1}{\alpha} \sum^{\alpha}_{f=1} \mathbb{I}(\tilde{s}_f \ge \tau) }
             {\frac{1}{\beta} \sum^{\beta}_{g=1} \mathbb{I}(s_g \ge \tau)},
\end{equation}
where $\tau$ is a given objectness threshold, and $\mathbb{I}(\cdot)$ is the indicator function that returns 1 if the input predicate is true and 0 otherwise. Intuitively, NDR measures the proportion of frames where the object is detected, hence a more effective attack will yield a lower NDR.

% \vfill
% \pagebreak
%%%%%%%%%%%%%%%%%%%%%%%%%%%%%%%%%%%%%%%%%%%%%%%%%%%%%%%%%%%%%%%%%%%%%%%%%%%%%%%%%%%%%%%%%
\section{Results}\label{sec:results}

We first perform ablation studies on the augmentations in Sec.~\ref{sec:patch} by evaluating the efficacy of $P^\ast$ in the digital domain (see Sec.~\ref{sec:aorr}). Then we evaluate the efficacy of $P^\ast$ in the physical domain (see Sec.~\ref{sec:osr_ndr}). 

\subsection{Results for digital domain evaluation}\label{sec:eval_digital}

\begin{figure}
    \centering
    \begin{subfigure}[b]{0.23\textwidth}
        \centering
        \includegraphics[width=\textwidth]{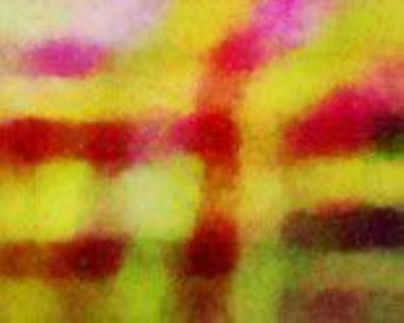}
        \caption{Type ON G/C+W optimised on Car Park.}
        \label{fig:patch04GCW}
    \end{subfigure}
    \hfill
    \begin{subfigure}[b]{0.23\textwidth}
        \centering
        \includegraphics[width=\textwidth]{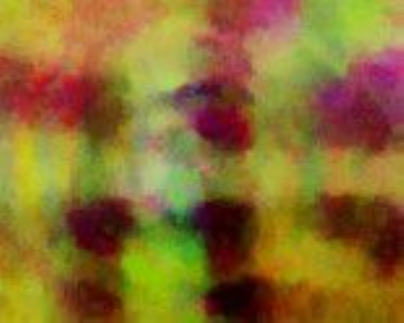}
        \caption{Type ON G/C+W optimised on Side Street.}
        \label{fig:patch12GCW}
    \end{subfigure}
    \hfill
    \begin{subfigure}[b]{0.23\textwidth}
        \centering
        \includegraphics[width=\textwidth]{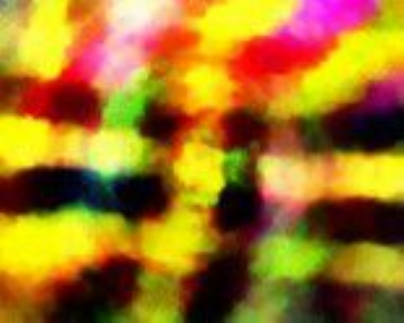}
        \caption{Type ON G/C optimised on Car Park.}
        \label{fig:patch04GC}
    \end{subfigure}
    \hfill
    \begin{subfigure}[b]{0.23\textwidth}
        \centering
        \includegraphics[width=\textwidth]{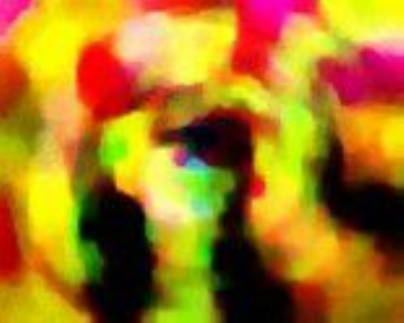}
        \caption{Type ON G/C optimised on Side Street.}
        \label{fig:patch12GC}
    \end{subfigure}
    \hfill
    \begin{subfigure}[b]{0.48\textwidth}
        \centering
        \includegraphics[width=\textwidth]{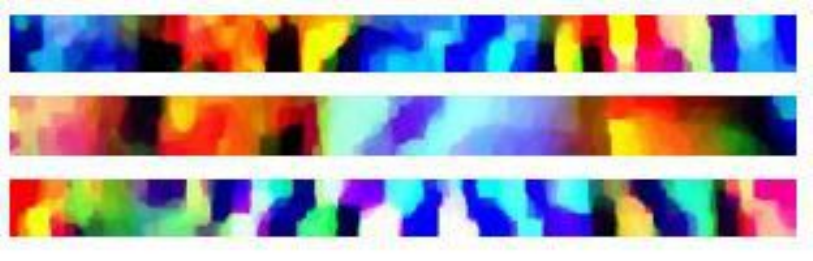}
        \caption{Type OFF G/C optimised on Side Street.}
        \label{fig:patch6CGC}
    \end{subfigure}
    \caption{Optimised patches $P^\ast$ in our experiments.}
    \label{fig:patches}
\end{figure}

Based on the collected data (Side Street and Car Park; see Sec.~\ref{sec:data-collection}), we optimised adversarial patches under different variants of the pipeline (Sec.~\ref{sec:patch}):
\begin{itemize}[leftmargin=1em,itemsep=2pt,parsep=0pt,topsep=2pt]
    \item G/C+W: geometric, colour-space and weather augmentations were applied (full pipeline).
    \item G/C: only geometric and colour-space augmentations.
    \item Control: patches were random intensity patterns, \ie, the optimisation pipeline was completely bypassed.
\end{itemize}
The weights for the terms in~\eqref{eq:loss} was $\delta = 0.01$ and $\gamma = 2.5$. Type OFF patch was not optimised for Car Park, due to the close proximity of the cars parked in the scene which prevented the off car patch from being embedded without occluding cars. Fig.~\ref{fig:patches} depicts the resulting patches. Note that G/C+W patches appear dimmer than G/C patches, suggesting that the optimisation is accounting for changes in scene appearance due to sun brightness and weather (the practical usefulness of this will be discussed below). 

\begin{figure}
     \centering
     \begin{subfigure}[b]{0.5\textwidth}
         \centering
         \includegraphics[width=\textwidth]{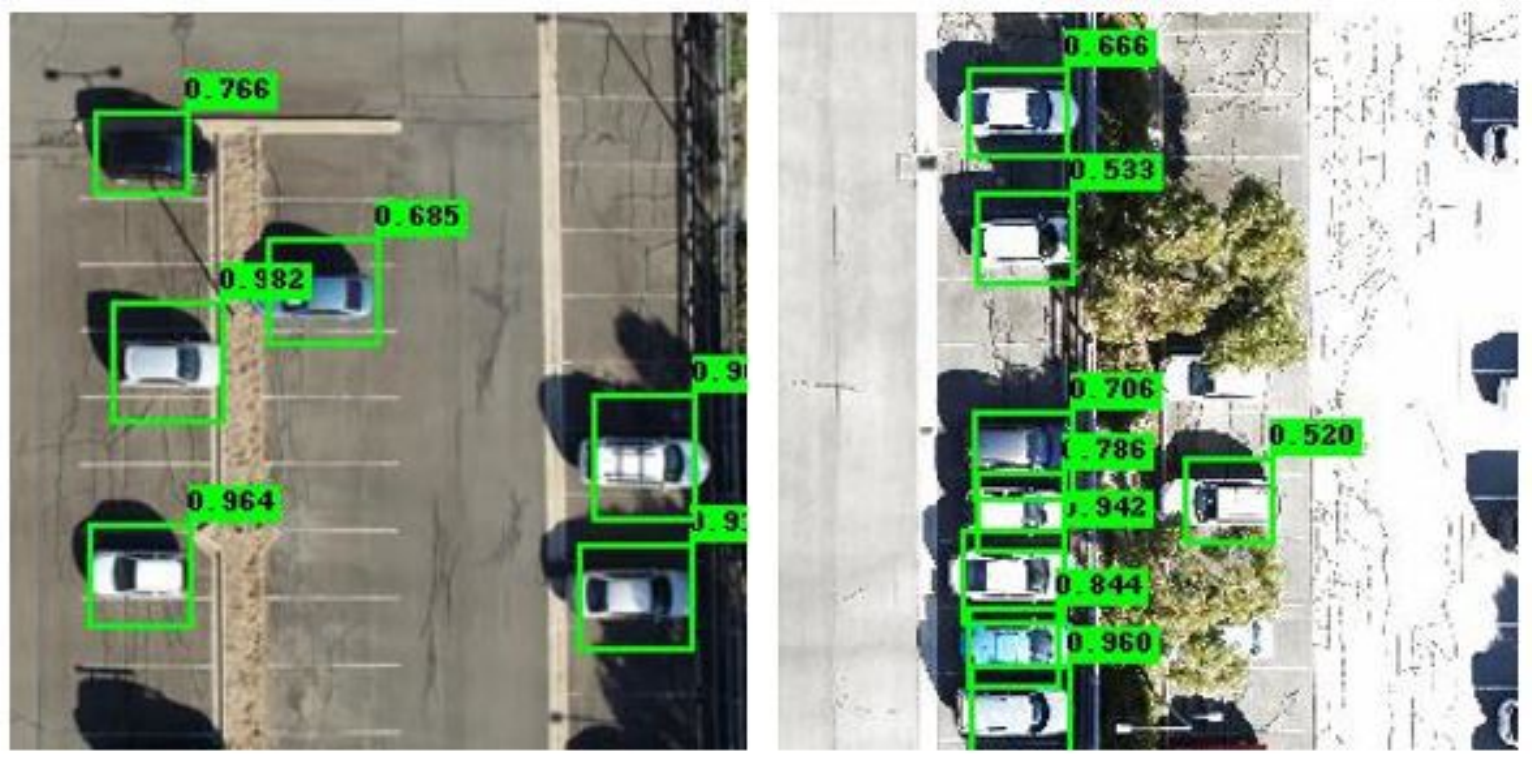}
         \caption{No patches embedded in an original testing image (STD) and the weather (snow) augmented version of the testing image (STD-W).}
         \label{fig:carpark_clean}
     \end{subfigure}
     \hfill
     \begin{subfigure}[b]{0.5\textwidth}
         \centering
         \includegraphics[width=\textwidth]{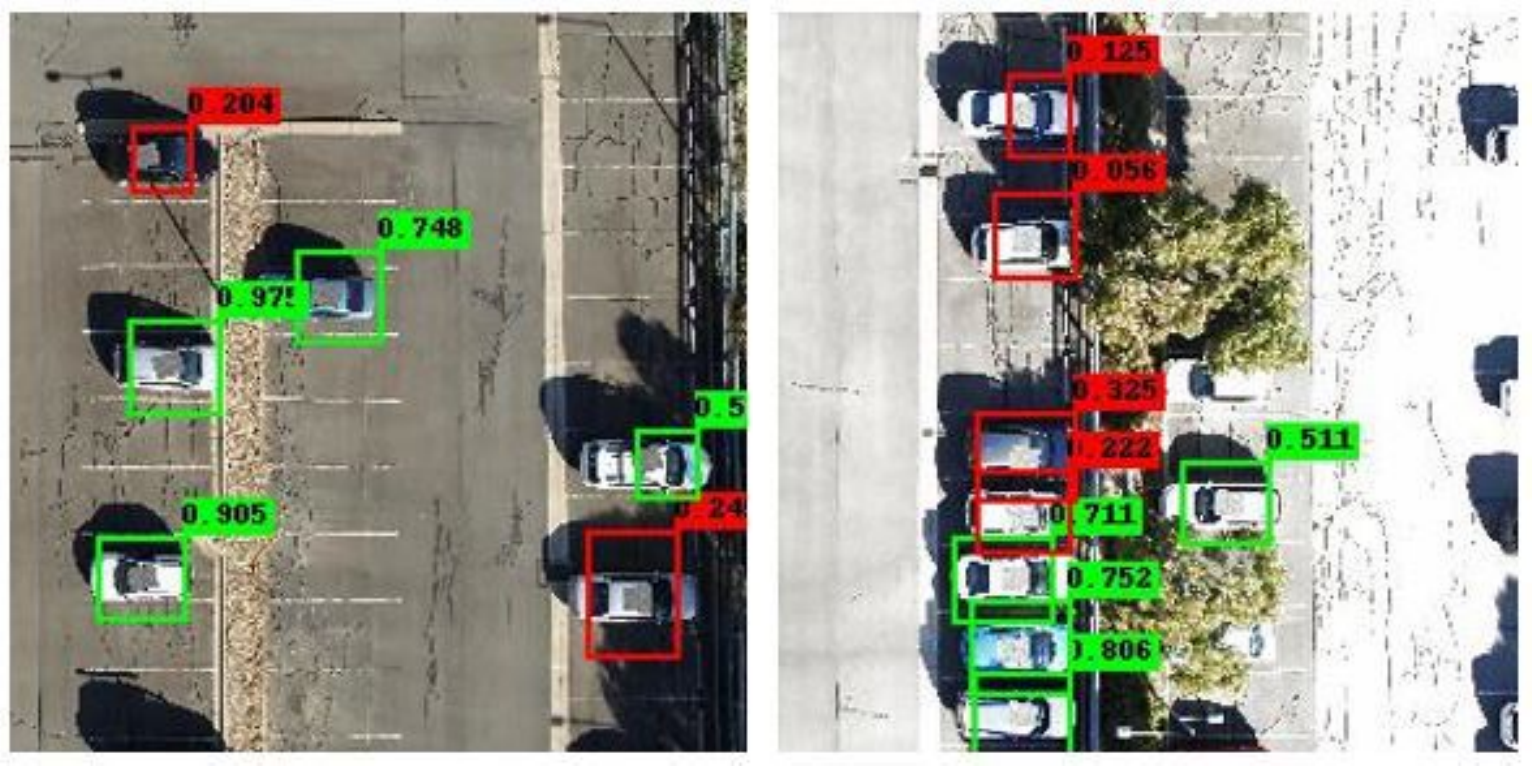}
         \caption{Embedded with Control patches (random intensity patterns).}
         \label{fig:carpark_controlpatch}
     \end{subfigure}     
     \hfill
     \begin{subfigure}[b]{0.5\textwidth}
         \centering
         \includegraphics[width=\textwidth]{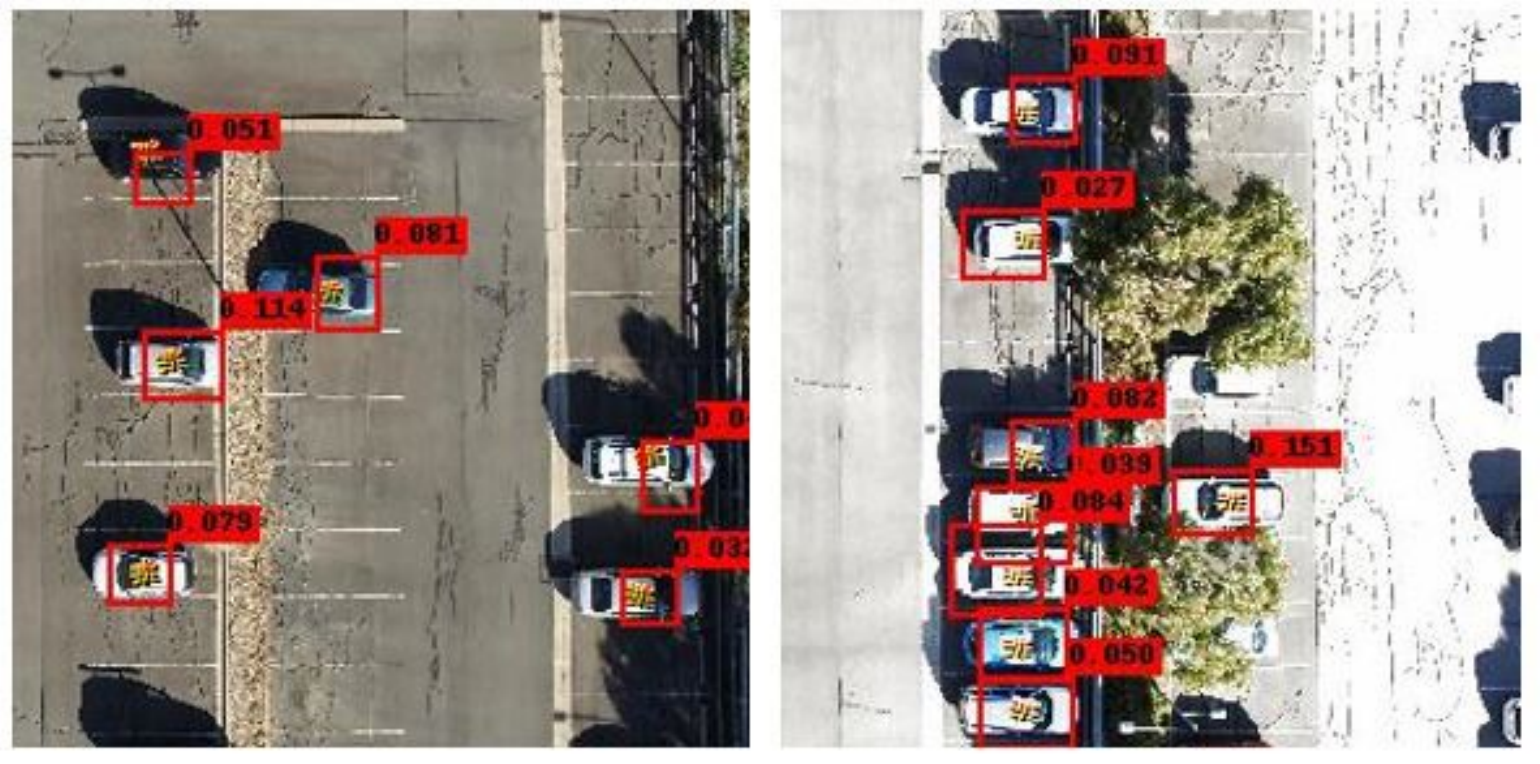}
         \caption{Embedded with patches optimised with the G/C pipeline.}
         \label{fig:carpark_gcpatch}
     \end{subfigure}
        \caption{Qualitative results from digital domain evaluation in the Car Park scene. Green and red bounding boxes indicate objectness scores $\ge 0.5$ and $<0.5$ respectively.}
        \label{fig:digital_attack}
\end{figure}

% Explain what the other columns of the table mean
Two variants of the testing regime (Sec.~\ref{sec:aorr}) were used:
\begin{itemize}[leftmargin=1em,itemsep=2pt,parsep=0pt,topsep=2pt]
    \item STD: following Sec.~\ref{sec:aorr} exactly.
    \item STD-W: the testing images $\cV$ were augmented with weather effects using the same steps in Sec.~\ref{sec:patch}.
\end{itemize}
Fig.~\ref{fig:digital_attack} illustrates sample qualitative results from our digital domain evaluation, while Table~\ref{tab:aorr_results} shows quantitative results.

\begin{table}[h]
    \setlength\tabcolsep{1pt}
    \begin{subtable}[h]{0.45\textwidth}
        \centering
        \begin{tabular}{l | c | c | c}
        Training pipeline     & Patch type         & AORR (STD) & AORR (STD+W) \\
        \hline \hline
        G/C+W & ON        & 0.602           & 0.558 \\
        G/C         & ON        & 0.712           & 0.615 \\
        Control       & ON        & 0.290           & 0.303 \\
        \hline
        G/C+W         & OFF       & -  & - \\
        G/C         & OFF       & 0.814  & 0.706 \\
        Control       & OFF       & 0.220           & 0.291 \\
        \end{tabular}
        \caption{Side Street.}
        % In total, 141 and 108 cars were attacked in the test set with and without weather effects respectively.
        \label{tab:aorr_sidestreet}
     \end{subtable}
     \begin{subtable}[h]{0.45\textwidth}
        \centering
        \begin{tabular}{l | c | c | c}
        Training pipeline     & Patch type         & AORR (STD) & AORR (STD+W) \\
        \hline\hline
        G/C+W & ON        & 0.856         & 0.693 \\
        G/C         & ON        & 0.878  & 0.768 \\
        Control       & ON        & 0.513           & 0.294 \\
       \end{tabular}
       \caption{Car Park.}
    %   In total, 752 and 464 cars were attacked in the test set with and without weather effects respectively.
       \label{tab:aorr_carpark}
     \end{subtable}
     \caption{Efficacy of adversarial patches on Side Street and Car Park scenes of attack under digital domain evaluation (higher AORR implies more effective attack).}
     \label{tab:aorr_results}
\end{table}

Qualitative results suggest that while both weather effects alone and Control patches can reduce the objectness scores, optimised patches are more successful. This is confirmed by the AORR values in Table~\ref{tab:aorr_results}. While both G/C and G/C+W were significantly more effective than Control, G/C visibly outperforms G/C+W, which indicates the lack of value in performing weather augmentations during training; this finding also motivated us to ignore G/C+W for optimising Type OFF patches for Side Street. However, the results show that, for the same scene (Side Street), Type OFF G/C patches outperformed Type ON G/C patches.

\begin{figure*}[t]\centering
     \begin{subfigure}[b]{0.32\textwidth}
         \includegraphics[width=\textwidth]{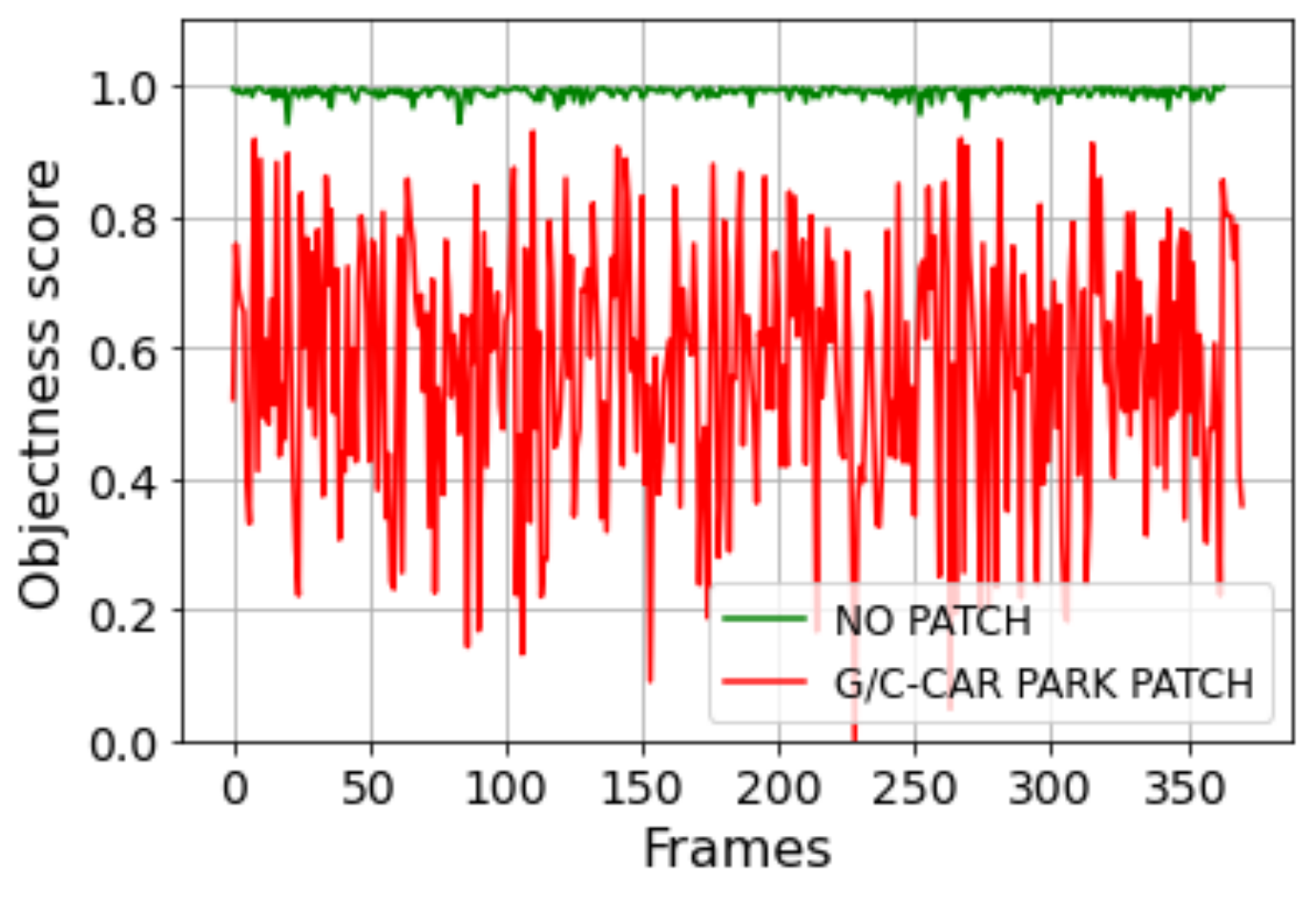}
         \caption{}
         \label{fig:freq_grey_carpark}
     \end{subfigure}
     \begin{subfigure}[b]{0.32\textwidth}
         \includegraphics[width=\textwidth]{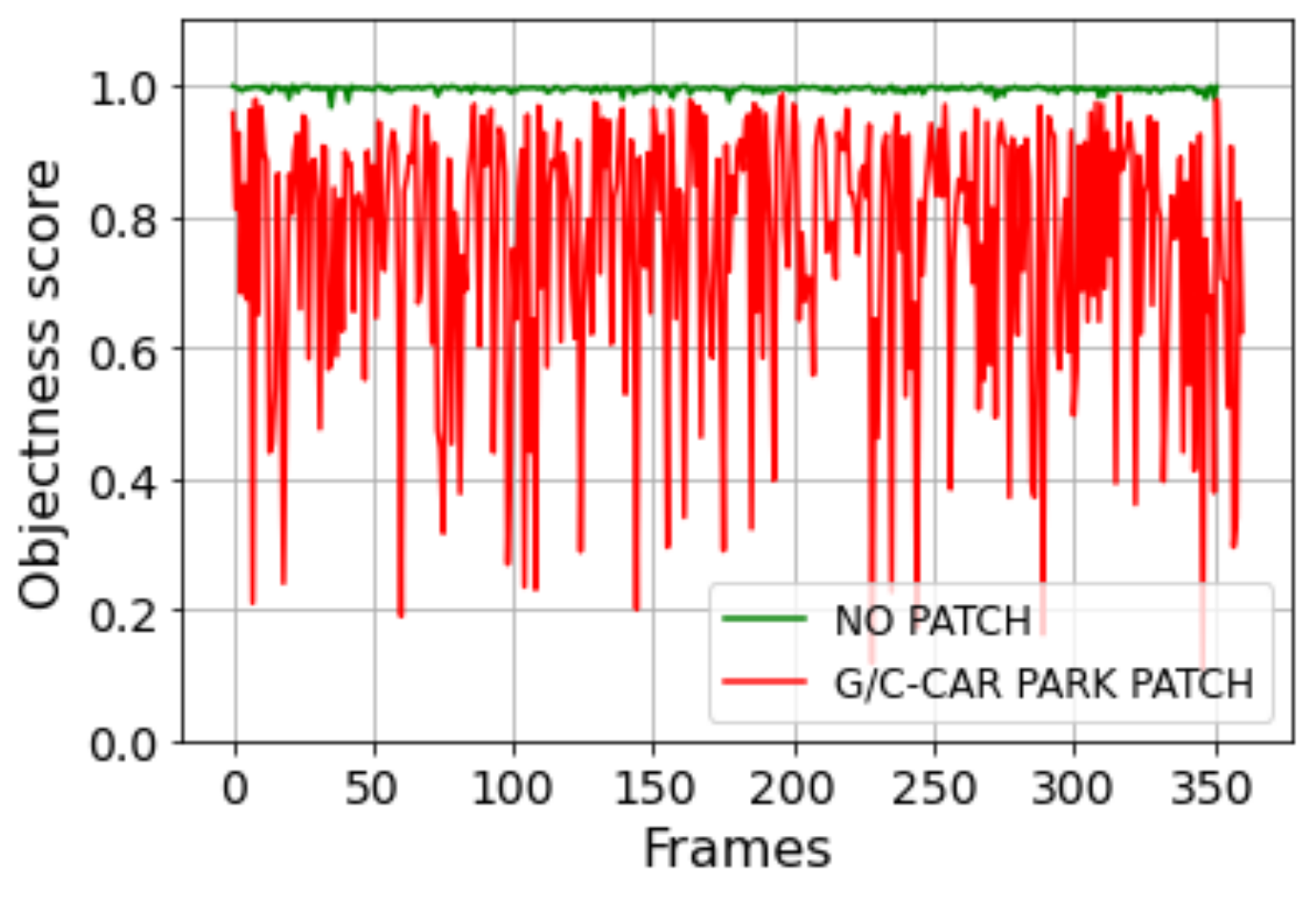}
         \caption{}
         \label{fig:freq_white_carpark}
     \end{subfigure}
     \begin{subfigure}[b]{0.32\textwidth}
         \includegraphics[width=\textwidth]{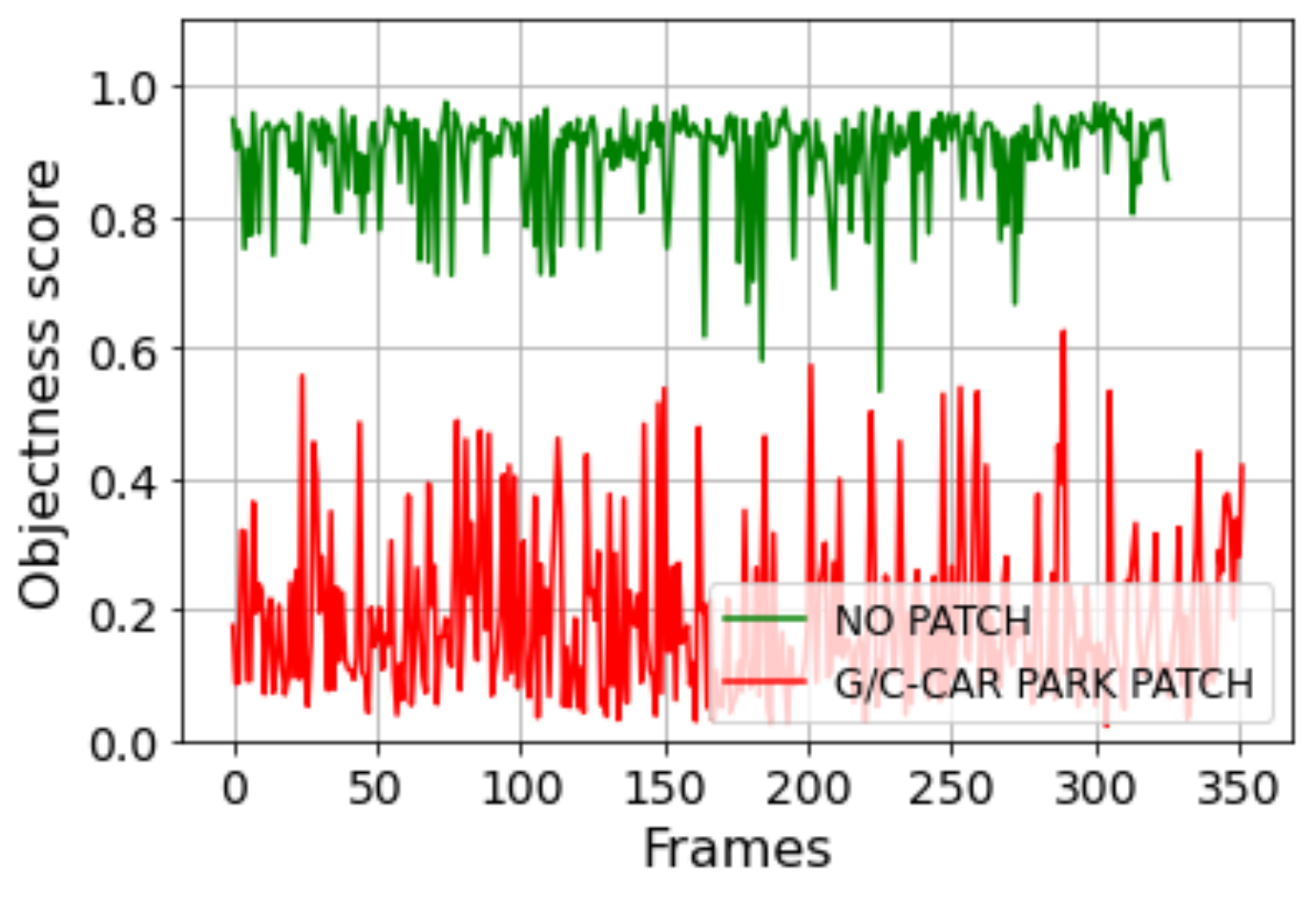}
         \caption{}
         \label{fig:freq_blue_carpark}
     \end{subfigure}
        \caption{Objectness scores for Car Park physical domain testing sets $\cF$ (with Type ON optimised patches $P^\ast$ installed in the cars) and $\cG$, plotted according to frame index, for the targeted (a) Grey, (b) White and (c) Blue cars.}
        \label{fig:variations}
\end{figure*}

\subsection{Results for physical domain evaluation}\label{sec:eval_physical}

Under the testing regime in Sec.~\ref{sec:osr_ndr}, patches were printed on 160 gsm coated paper, 25 FPS videos were captured to form the new testing images $\cF$ and $\cG$ for both Car Park and Side Street. When capturing $\cF$, three cars under our control (``Grey'', ``White'', ``Blue'') were installed with the physical realisations of the optimised patches $P^\ast$. Basic statistics of the data are as follows:
\begin{itemize}[leftmargin=1em,itemsep=2pt,parsep=0pt,topsep=2pt]
    \item Car Park: $|\cF| = 1,084$ frames, $|\cG| = 1,042$.
    \item Side Street: $|\cF| = 4,699$ frames, $|\cG| = 526$.
\end{itemize}
We further categorise the images into different settings:
\begin{itemize}[leftmargin=1em,itemsep=2pt,parsep=0pt,topsep=2pt]
    \item Lighting: whether the car was in sunlight or shade.
    \item Motion: whether the car was static or moving wrt ground.
\end{itemize}
Due to the cost of performing physical experiments (\eg, civil aviation approval, availability of personnel) and uncontrollable environmental factors, not all the combinations above were explored. Also, only patches optimised under the G/C setting were used based on the findings in Sec.~\ref{sec:data-collection}.

\begin{figure}[ht]
\centering\includegraphics[width=\linewidth]{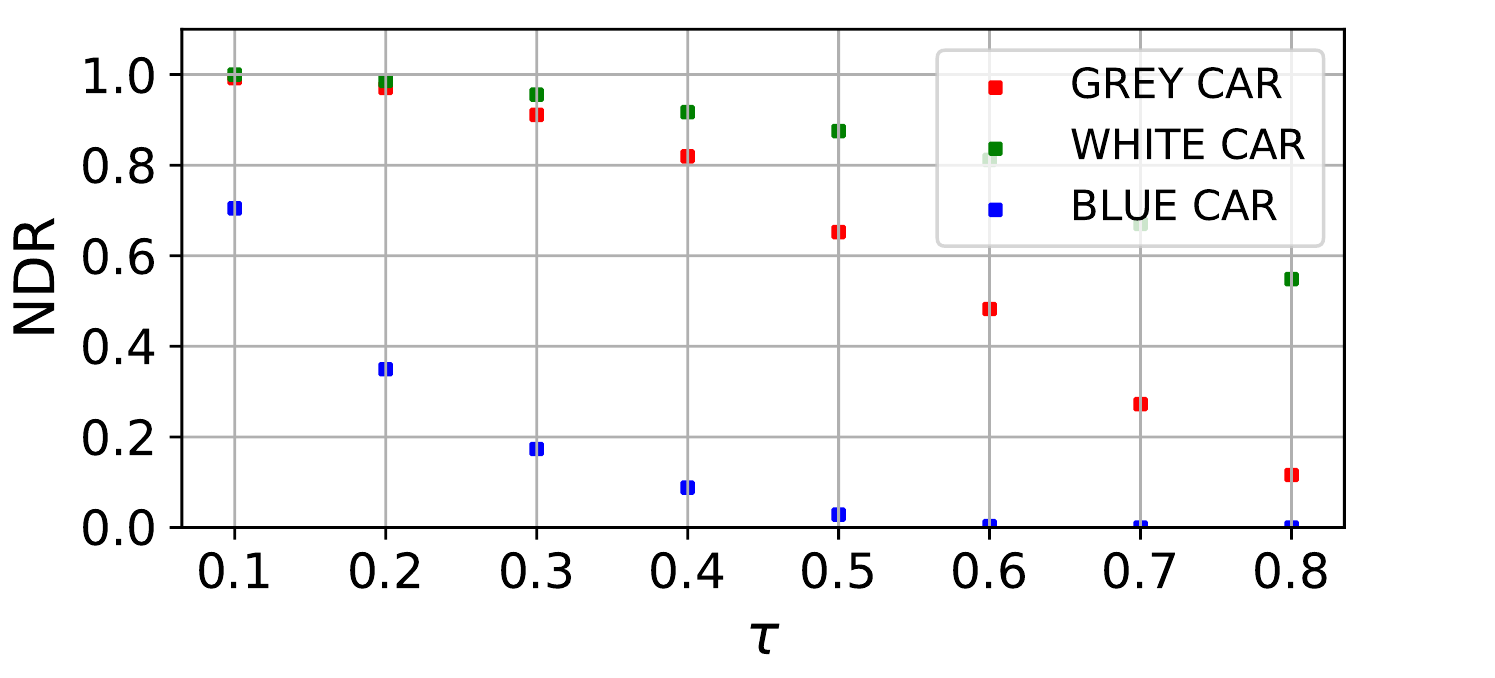}
\vspace{-2em}
\caption{Mean NDR as a function of objectness threshold $\tau$ for Car Park and Side Street (Type ON patches only).}
\label{fig:ncr_carpark}
\end{figure}

Qualitative results are illustrated in Fig.~\ref{fig:physical_attack}. Fig.~\ref{fig:variations} shows the objectness scores in $\cF$ and $\cG$ for the Car Park scene, which highlights differences in attack efficacy for the different cars. In particular, the attack was much more successful for Blue, even accounting for a visibly lower objectness score prior to the attack. Quantitative results (mean OSR; lower means more successful attack) in Table~\ref{tab:osr_results} illustrate the general effectiveness of the physical attack, \ie, significant reductions in objectness scores (25\% to 85\%) were achievable (depending on the car and environmental factors).  While Type OFF patches are successful in reducing objectness scores, in contrast to the digital evaluation they are less effective than Type ON patches. Fig.~\ref{fig:ncr_carpark} plots mean NDR as a function of objectness threshold $\tau$ (lower NDR means higher attack effectiveness) for the three cars, which also illustrates differences in attack efficacy.

\begin{table}[ht]
    \setlength\tabcolsep{1pt}
    \begin{subtable}[h]{0.45\textwidth}
        \centering
        \begin{tabular}{c | c | c | c | c}
        Car     & Patch type      & Lighting & Motion & Mean OSR   \\
        \hline\hline
        Grey    & ON     & Both      & Moving         & 0.343 \\
        Grey    & ON     & Sun       & Static         & 0.251 \\
        Grey    & ON     & Shade     & Static         & 0.255 \\
        \hline
        Grey    & OFF    & Sun       & Static         & 0.429 \\
        \hline
        \hline
        White   & ON     & Both      & Moving         & 0.286 \\
        White   & ON     & Sun       & Static         & 0.285 \\
        White   & ON     & Shade     & Static         & 0.197 \\
        \hline
        White   & OFF    & Sun       & Static         & 0.748 \\
        \end{tabular}
        \caption{Side Street.}
        \label{tab:osr_sidestreet}
    \end{subtable}
    \hfill
    \begin{subtable}[h]{0.45\textwidth}
        \centering
        \begin{tabular}{c | c | c | c | c}
        Car     & Patch type  & Lighting & Motion & Mean OSR   \\ 
        \hline\hline
        Grey    & ON     & Sun       & Static        & 0.509 \\
        Blue    & ON     & Shade     & Static        & 0.208 \\
        White   & ON     & Sun       & Static        & 0.746 \\
       \end{tabular}
       \caption{Car Park.}
       \label{tab:osr_carpark}
    \end{subtable}
    \caption{Mean OSR in Car Park and Side Street.}
    \label{tab:osr_results}
\end{table}

%%%%%%%%%%%%%%%%%%%%%%%%%%%%%%%%%%%%%%%%%%%%%%%%%%%%%%%%%%%%%%%%%%%%%%%%%%%%%%%%%%%%%%%%%%
\section{Conclusions}\label{sec:conclusion}

We demonstrated physical adversarial attacks on a car detector in aerial scenes, and proposed a novel ``off car'' patch design which was shown to be effective. Our results indicate that, while the efficacy of the attack is subject to atmospheric factors (lighting, weather, seasons) and the target-observer distance, physical adversarial attacks are a realistic threat. Curiously, our ablation tests showed that augmenting patch optimisation with weather effects did not result in higher effectiveness, even in the digital domain. This will form a useful topic for future investigations.

\section*{Acknowledgements}\label{sec:acknowledgement}
Tat-Jun Chin is SmartSat CRC Professorial Chair of Sentient Satellites.

%\hfill\break
% \clearpage
% \newpage

%%%%%%%%%%%%%%%%%%%%%%%%%%%%%%%%%%%%%%%%%%%%%%%%%%%%%%%%%%%%%%%%%%%%%%%%%%%%%%%%%%%%%%%%%%
{\small
\bibliographystyle{ieee_fullname}
\bibliography{egbib}

\begin{thebibliography}{10}\itemsep=-1pt

\bibitem{akhtar2018threat}
Naveed Akhtar and Ajmal Mian.
\newblock Threat of adversarial attacks on deep learning in computer vision: A
  survey.
\newblock {\em IEEE Access}, 6:14410--14430, 2018.

\bibitem{albert2017using}
Adrian Albert, Jasleen Kaur, and Marta~C. Gonzalez.
\newblock Using convolutional networks and satellite imagery to identify
  patterns in urban environments at a large scale.
\newblock In {\em Proceedings of the 23rd ACM SIGKDD International Conference
  on Knowledge Discovery and Data Mining}, KDD '17, page 1357–1366, New York,
  NY, USA, 2017. Association for Computing Machinery.

\bibitem{athalye2017synthesizing}
Anish Athalye, Logan Engstrom, Andrew Ilyas, and Kevin Kwok.
\newblock Synthesizing robust adversarial examples.
\newblock {\em arXiv preprint arXiv:1707.07397}, 2017.

\bibitem{brendel2018decision}
Wieland Brendel, Jonas Rauber, and Matthias Bethge.
\newblock Decision-based adversarial attacks: Reliable attacks against
  black-box machine learning models.
\newblock In {\em International Conference on Learning Representations}, 2018.

\bibitem{brown2017adversarial}
Tom~B Brown, Dandelion Man{\'e}, Aurko Roy, Mart{\'\i}n Abadi, and Justin
  Gilmer.
\newblock Adversarial patch.
\newblock In {\em 31st Conference on Neural Information Processing Systems
  (NIPS 2017)}, 2017.

\bibitem{carlini2017magnet}
Nicholas Carlini and David Wagner.
\newblock {MagNet} and ``efficient defenses against adversarial attacks'' are
  not robust to adversarial examples.
\newblock {\em arXiv preprint arXiv:1711.08478}, 2017.

\bibitem{carlini2017towards}
Nicholas Carlini and David Wagner.
\newblock Towards evaluating the robustness of neural networks.
\newblock In {\em 2017 IEEE Symposium on Security and Privacy (SP)}, pages
  39--57. IEEE, 2017.

\bibitem{chen2018shapeshifter}
Shang-Tse Chen, Cory Cornelius, Jason Martin, and Duen Horng~(Polo) Chau.
\newblock {ShapeShifter}: Robust physical adversarial attack on {Faster R-CNN}
  object detector.
\newblock In Michele Berlingerio, Francesco Bonchi, Thomas G{\"a}rtner, Neil
  Hurley, and Georgiana Ifrim, editors, {\em Machine Learning and Knowledge
  Discovery in Databases}, pages 52--68, Cham, 2019. Springer International
  Publishing.

\bibitem{czaja2018adversarial}
Wojciech Czaja, Neil Fendley, Michael Pekala, Christopher Ratto, and I-Jeng
  Wang.
\newblock Adversarial examples in remote sensing.
\newblock In {\em Proceedings of the 26th ACM SIGSPATIAL International
  Conference on Advances in Geographic Information Systems}, SIGSPATIAL '18,
  page 408–411, New York, NY, USA, 2018. Association for Computing Machinery.

\bibitem{den2020adversarial}
Richard den Hollander, Ajaya Adhikari, Ioannis Tolios, Michael van Bekkum,
  Anneloes Bal, Stijn Hendriks, Maarten Kruithof, Dennis Gross, Nils Jansen,
  Guillermo Perez, Kit Buurman, and Stephan Raaijmakers.
\newblock {Adversarial patch camouflage against aerial detection}.
\newblock In Judith Dijk, editor, {\em Artificial Intelligence and Machine
  Learning in Defense Applications II}, volume 11543, pages 77--86.
  International Society for Optics and Photonics, SPIE, 2020.

\bibitem{deng2019arcface}
Jiankang Deng, Jia Guo, Niannan Xue, and Stefanos Zafeiriou.
\newblock {ArcFace}: Additive angular margin loss for deep face recognition.
\newblock In {\em Proceedings of the IEEE/CVF Conference on Computer Vision and
  Pattern Recognition (CVPR)}, pages 4690--4699, 2019.

\bibitem{du2021physical}
Andrew Du.
\newblock Physical adversarial attacks on an aerial imagery object detector.
\newblock GitHub repository at
  \url{https://github.com/andrewpatrickdu/adversarial-yolov3-cowc}, 2021.
\newblock Accessed 11 Oct 2021.

\bibitem{eykholt2018robust}
Kevin Eykholt, Ivan Evtimov, Earlence Fernandes, Tadayoshi Kohno, Bo Li, Atul
  Prakash, Amir Rahmati, and Dawn Song.
\newblock Robust physical-world attacks on deep learning models.
\newblock In {\em Computer Vision and Pattern Recognition}, 2018.

\bibitem{eykholt2017note}
Kevin Eykholt, Ivan Evtimov, Earlence Fernandes, Bo Li, Dawn Song, Tadayoshi
  Kohno, Amir Rahmati, Atul Prakash, and Florian Tramer.
\newblock Note on attacking object detectors with adversarial stickers.
\newblock {\em arXiv preprint arXiv:1712.08062}, 2017.

\bibitem{goodfellow2015explaining}
Ian Goodfellow, Jonathon Shlens, and Christian Szegedy.
\newblock Explaining and harnessing adversarial examples.
\newblock In {\em International Conference on Learning Representations}, 2015.

\bibitem{gu2014towards}
Shixiang Gu and Luca Rigazio.
\newblock Towards deep neural network architectures robust to adversarial
  examples.
\newblock {\em arXiv preprint arXiv:1412.5068}, 2014.

\bibitem{huang2020universal}
Lifeng Huang, Chengying Gao, Yuyin Zhou, Cihang Xie, Alan~L Yuille, Changqing
  Zou, and Ning Liu.
\newblock Universal physical camouflage attacks on object detectors.
\newblock In {\em Proceedings of the IEEE/CVF Conference on Computer Vision and
  Pattern Recognition}, pages 720--729, 2020.

\bibitem{cvat}
Intel.
\newblock {Computer Vision Annotation Tool (CVAT)}.
\newblock GitHub repository, 2021.
\newblock v1.6.0, accessed 6 October 2021.

\bibitem{isola2017image}
Phillip Isola, Jun-Yan Zhu, Tinghui Zhou, and Alexei~A. Efros.
\newblock Image-to-image translation with conditional adversarial networks.
\newblock In {\em Proceedings of the IEEE Conference on Computer Vision and
  Pattern Recognition (CVPR)}, July 2017.

\bibitem{jan2019connecting}
Steve~T.K. Jan, Joseph Messou, Yen-Chen Lin, Jia-Bin Huang, and Gang Wang.
\newblock Connecting the digital and physical world: Improving the robustness
  of adversarial attacks.
\newblock {\em Proceedings of the AAAI Conference on Artificial Intelligence},
  33(01):962--969, July 2019.

\bibitem{ji2020vehicle}
Hong Ji, Zhi Gao, Tiancan Mei, and Bharath Ramesh.
\newblock Vehicle detection in remote sensing images leveraging on simultaneous
  super-resolution.
\newblock {\em IEEE Geoscience and Remote Sensing Letters}, 17(4):676--680,
  2020.

\bibitem{kannan2018adversarial}
Harini Kannan, Alexey Kurakin, and Ian Goodfellow.
\newblock Adversarial logit pairing.
\newblock {\em arXiv preprint arXiv:1803.06373}, 2018.

\bibitem{kingma2014adam}
Diederik~P Kingma and Jimmy Ba.
\newblock Adam: A method for stochastic optimization.
\newblock {\em arXiv preprint arXiv:1412.6980}, 2014.

\bibitem{komkov2021advhat}
Stepan Komkov and Aleksandr Petiushko.
\newblock {AdvHat: Real-World Adversarial Attack on ArcFace Face ID System}.
\newblock In {\em 2020 25th International Conference on Pattern Recognition
  (ICPR)}, pages 819--826, 2021.

\bibitem{kurakin2016adversarial}
Alexey Kurakin, Ian Goodfellow, and Samy Bengio.
\newblock Adversarial examples in the physical world.
\newblock {\em arXiv preprint arXiv:1607.02533}, 2016.

\bibitem{kurakin2017adversarial}
Alexey Kurakin, Ian~J. Goodfellow, and Samy Bengio.
\newblock Adversarial machine learning at scale.
\newblock In {\em International Conference on Learning Representations (ICLR)},
  2017.

\bibitem{lee2019physical}
Mark Lee and Zico Kolter.
\newblock On physical adversarial patches for object detection.
\newblock In {\em ICML 2019 Workshop on the Security and Privacy of Machine
  Learning}, 2019.

\bibitem{lin2014microsoft}
Tsung-Yi Lin, Michael Maire, Serge Belongie, James Hays, Pietro Perona, Deva
  Ramanan, Piotr Doll{\'a}r, and C~Lawrence Zitnick.
\newblock Microsoft coco: Common objects in context.
\newblock In {\em European conference on computer vision}, pages 740--755.
  Springer, 2014.

\bibitem{lu2017adversarial}
Jiajun Lu, Hussein Sibai, and Evan Fabry.
\newblock Adversarial examples that fool detectors.
\newblock {\em arXiv preprint arXiv:1712.02494}, 2017.

\bibitem{lu2017no}
Jiajun Lu, Hussein Sibai, Evan Fabry, and David Forsyth.
\newblock No need to worry about adversarial examples in object detection in
  autonomous vehicles.
\newblock {\em arXiv preprint arXiv:1707.03501}, 2017.

\bibitem{madry2017towards}
Aleksander Madry, Aleksandar Makelov, Ludwig Schmidt, Dimitris Tsipras, and
  Adrian Vladu.
\newblock Towards deep learning models resistant to adversarial attacks.
\newblock {\em arXiv preprint arXiv:1706.06083}, 2017.

\bibitem{manfreda2018use}
Salvatore Manfreda, Matthew~F. McCabe, Pauline~E. Miller, Richard Lucas, Victor
  Pajuelo~Madrigal, Giorgos Mallinis, Eyal Ben~Dor, David Helman, Lyndon Estes,
  Giuseppe Ciraolo, Jana Müllerová, Flavia Tauro, M.~Isabel De~Lima, João L.
  M.~P. De~Lima, Antonino Maltese, Felix Frances, Kelly Caylor, Marko Kohv,
  Matthew Perks, Guiomar Ruiz-Pérez, Zhongbo Su, Giulia Vico, and Brigitta
  Toth.
\newblock On the use of unmanned aerial systems for environmental monitoring.
\newblock {\em Remote Sensing}, 10(4), 2018.

\bibitem{moosavi2017universal}
Seyed-Mohsen Moosavi-Dezfooli, Alhussein Fawzi, Omar Fawzi, and Pascal
  Frossard.
\newblock Universal adversarial perturbations.
\newblock In {\em Proceedings of the IEEE conference on computer vision and
  pattern recognition}, pages 1765--1773, 2017.

\bibitem{moosavi2016deepfool}
Seyed-Mohsen Moosavi-Dezfooli, Alhussein Fawzi, and Pascal Frossard.
\newblock Deepfool: a simple and accurate method to fool deep neural networks.
\newblock In {\em Proceedings of the IEEE Conference on Computer Vision and
  Pattern Recognition}, pages 2574--2582, 2016.

\bibitem{mundhenk2016large}
T~Nathan Mundhenk, Goran Konjevod, Wesam~A Sakla, and Kofi Boakye.
\newblock A large contextual dataset for classification, detection and counting
  of cars with deep learning.
\newblock In {\em European Conference on Computer Vision}, pages 785--800.
  Springer, 2016.

\bibitem{papernot2016transferability}
Nicolas Papernot, Patrick McDaniel, and Ian Goodfellow.
\newblock Transferability in machine learning: from phenomena to black-box
  attacks using adversarial samples.
\newblock {\em arXiv preprint arXiv:1605.07277}, 2016.

\bibitem{papernot2017practical}
Nicolas Papernot, Patrick McDaniel, Ian Goodfellow, Somesh Jha, Z~Berkay Celik,
  and Ananthram Swami.
\newblock Practical black-box attacks against machine learning.
\newblock In {\em Proceedings of the 2017 ACM on Asia Conference on Computer
  and Communications Security}, pages 506--519. ACM, 2017.

\bibitem{papernot2016limitations}
Nicolas Papernot, Patrick McDaniel, Somesh Jha, Matt Fredrikson, Z~Berkay
  Celik, and Ananthram Swami.
\newblock The limitations of deep learning in adversarial settings.
\newblock In {\em 2016 IEEE European Symposium on Security and Privacy
  (EuroS\&P)}, pages 372--387. IEEE, 2016.

\bibitem{papernot2016distillation}
Nicolas Papernot, Patrick McDaniel, Xi Wu, Somesh Jha, and Ananthram Swami.
\newblock Distillation as a defense to adversarial perturbations against deep
  neural networks.
\newblock In {\em 2016 IEEE Symposium on Security and Privacy (SP)}, pages
  582--597. IEEE, 2016.

\bibitem{pritt2017satellite}
Mark Pritt and Gary Chern.
\newblock Satellite image classification with deep learning.
\newblock In {\em 2017 IEEE Applied Imagery Pattern Recognition Workshop
  (AIPR)}, pages 1--7, 2017.

\bibitem{redmon2016you}
Joseph Redmon, Santosh Divvala, Ross Girshick, and Ali Farhadi.
\newblock You only look once: Unified, real-time object detection.
\newblock In {\em Proceedings of the IEEE conference on computer vision and
  pattern recognition}, pages 779--788, 2016.

\bibitem{redmon2017yolo9000}
Joseph Redmon and Ali Farhadi.
\newblock {YOLO9000}: better, faster, stronger.
\newblock In {\em Proceedings of the IEEE conference on computer vision and
  pattern recognition}, pages 7263--7271, 2017.

\bibitem{redmon2018yolov3}
Joseph Redmon and Ali Farhadi.
\newblock {YOLOv3}: An incremental improvement.
\newblock {\em arXiv preprint arXiv:1804.02767}, 2018.

\bibitem{ren2015faster}
Shaoqing Ren, Kaiming He, Ross Girshick, and Jian Sun.
\newblock Faster {R-CNN}: Towards real-time object detection with region
  proposal networks.
\newblock In {\em Advances in neural information processing systems}, pages
  91--99, 2015.

\bibitem{automold2018}
Ujjwal Saxena.
\newblock Automold.
\newblock Road augmentation library at
  \href{https://github.com/UjjwalSaxena/Automold--Road-Augmentation-Library}{https://github.com/UjjwalSaxena/Automold\---Road-Augmentation-Library},
  2018.

\bibitem{sharif2016accessorize}
Mahmood Sharif, Sruti Bhagavatula, Lujo Bauer, and Michael~K Reiter.
\newblock Accessorize to a crime: Real and stealthy attacks on state-of-the-art
  face recognition.
\newblock In {\em Proceedings of the 2016 ACM SIGSAC Conference on Computer and
  Communications Security}, pages 1528--1540. ACM, 2016.

\bibitem{shermeyer2019effects}
Jacob Shermeyer and Adam Van~Etten.
\newblock The effects of super-resolution on object detection performance in
  satellite imagery.
\newblock In {\em Proceedings of the IEEE/CVF Conference on Computer Vision and
  Pattern Recognition (CVPR) Workshops}, June 2019.

\bibitem{sitawarin2018darts}
Chawin Sitawarin, Arjun~Nitin Bhagoji, Arsalan Mosenia, Mung Chiang, and
  Prateek Mittal.
\newblock Darts: Deceiving autonomous cars with toxic signs.
\newblock {\em arXiv preprint arXiv:1802.06430}, 2018.

\bibitem{sitawarin2018rogue}
Chawin Sitawarin, Arjun~Nitin Bhagoji, Arsalan Mosenia, Prateek Mittal, and
  Mung Chiang.
\newblock Rogue signs: Deceiving traffic sign recognition with malicious ads
  and logos.
\newblock {\em arXiv preprint arXiv:1801.02780}, 2018.

\bibitem{song2018physical}
Dawn Song, Kevin Eykholt, Ivan Evtimov, Earlence Fernandes, Bo Li, Amir
  Rahmati, Florian Tram{\`e}r, Atul Prakash, and Tadayoshi Kohno.
\newblock Physical adversarial examples for object detectors.
\newblock In {\em 12th {USENIX} Workshop on Offensive Technologies ({WOOT}
  18)}, Baltimore, MD, Aug. 2018. {USENIX} Association.

\bibitem{su2019one}
Jiawei Su, Danilo~Vasconcellos Vargas, and Kouichi Sakurai.
\newblock One pixel attack for fooling deep neural networks.
\newblock {\em IEEE Transactions on Evolutionary Computation}, 2019.

\bibitem{szegedy2016rethinking}
Christian Szegedy, Vincent Vanhoucke, Sergey Ioffe, Jon Shlens, and Zbigniew
  Wojna.
\newblock Rethinking the inception architecture for computer vision.
\newblock In {\em Proceedings of the IEEE conference on computer vision and
  pattern recognition}, pages 2818--2826, 2016.

\bibitem{szegedy2013intriguing}
Christian Szegedy, Wojciech Zaremba, Ilya Sutskever, Joan Bruna, Dumitru Erhan,
  Ian Goodfellow, and Rob Fergus.
\newblock Intriguing properties of neural networks.
\newblock {\em arXiv preprint arXiv:1312.6199}, 2013.

\bibitem{nistir8269}
Elham Tabassi, Kevin~J. Burns, Michael Hadjimichael, Andres~D. Molina-Markham,
  and Julian~T. Sexton.
\newblock A taxonomy and terminology of adversarial machine learning.
\newblock Draft NISTIR 8269, National Institution of Standards and Technology,
  2019.

\bibitem{thys2019fooling}
Simen Thys, Wiebe Van~Ranst, and Toon Goedeme.
\newblock Fooling automated surveillance cameras: Adversarial patches to attack
  person detection.
\newblock In {\em Proceedings of the IEEE/CVF Conference on Computer Vision and
  Pattern Recognition (CVPR) Workshops}, June 2019.

\bibitem{wang2016theoretical}
Beilun Wang, Ji Gao, and Yanjun Qi.
\newblock A theoretical framework for robustness of (deep) classifiers against
  adversarial examples.
\newblock {\em arXiv preprint arXiv:1612.00334}, 2016.

\bibitem{wang2021dual}
Jiakai Wang, Aishan Liu, Zixin Yin, Shunchang Liu, Shiyu Tang, and Xianglong
  Liu.
\newblock Dual attention suppression attack: Generate adversarial camouflage in
  physical world.
\newblock In {\em Proceedings of the IEEE/CVF Conference on Computer Vision and
  Pattern Recognition (CVPR)}, pages 8565--8574, June 2021.

\bibitem{wang2021towards}
Yajie Wang, Haoran Lv, Xiaohui Kuang, Gang Zhao, Yu-an Tan, Quanxin Zhang, and
  Jingjing Hu.
\newblock Towards a physical-world adversarial patch for blinding object
  detection models.
\newblock {\em Information Sciences}, 556:459--471, 2021.

\bibitem{wu2020making}
Zuxuan Wu, Ser-Nam Lim, Larry~S. Davis, and Tom Goldstein.
\newblock Making an invisibility cloak: Real world adversarial attacks on
  object detectors.
\newblock In Andrea Vedaldi, Horst Bischof, Thomas Brox, and Jan-Michael Frahm,
  editors, {\em Computer Vision -- ECCV 2020}, pages 1--17, Cham, 2020.
  Springer International Publishing.

\bibitem{xu2020adversarial}
Kaidi Xu, Gaoyuan Zhang, Sijia Liu, Quanfu Fan, Mengshu Sun, Hongge Chen,
  Pin-Yu Chen, Yanzhi Wang, and Xue Lin.
\newblock {Adversarial T-Shirt! Evading Person Detectors in a Physical World}.
\newblock In {\em Computer Vision -- ECCV 2020}, pages 665--681, Cham, 2020.
  Springer International Publishing.

\bibitem{xu2018feature}
Weilin Xu, David Evans, and Yanjun Qi.
\newblock Feature squeezing: Detecting adversarial examples in deep neural
  networks.
\newblock In {\em Network and Distributed Systems Security Symposium (NDSS)},
  2018.

\bibitem{xue20203d}
Mingfu Xue, Can He, Zhiyu Wu, Jian Wang, Zhe Liu, and Weiqiang Liu.
\newblock {3D} invisible cloak.
\newblock {\em arXiv preprint arXiv:2011.13705}, 2020.

\bibitem{xue2021naturalae}
Mingfu Xue, Chengxiang Yuan, Can He, Jian Wang, and Weiqiang Liu.
\newblock {NaturalAE}: Natural and robust physical adversarial examples for
  object detectors.
\newblock {\em Journal of Information Security and Applications}, 57:102694,
  2021.

\bibitem{yeh2020using}
Christopher Yeh, Anthony Perez, Anne Driscoll, George Azzari, Zhongyi Tang,
  David Lobell, Stefano Ermon, and Marshall Burke.
\newblock Using publicly available satellite imagery and deep learning to
  understand economic well-being in {Africa}.
\newblock {\em Nature Communications}, 11(1):2583, May 2020.

\bibitem{yuan2019adversarial}
X. {Yuan}, P. {He}, Q. {Zhu}, and X. {Li}.
\newblock Adversarial examples: Attacks and defenses for deep learning.
\newblock {\em IEEE Transactions on Neural Networks and Learning Systems},
  30(9):2805--2824, 2019.

\bibitem{yuan2021review}
Xiaohui Yuan, Jianfang Shi, and Lichuan Gu.
\newblock A review of deep learning methods for semantic segmentation of remote
  sensing imagery.
\newblock {\em Expert Systems with Applications}, 169:114417, 2021.

\bibitem{zhang2019camou}
Yang Zhang, Hassan Foroosh, Philip David, and Boqing Gong.
\newblock {CAMOU}: Learning physical vehicle camouflages to adversarially
  attack detectors in the wild.
\newblock In {\em International Conference on Learning Representations}, 2019.

\bibitem{zhu2017unpaired}
Jun-Yan Zhu, Taesung Park, Phillip Isola, and Alexei~A Efros.
\newblock Unpaired image-to-image translation using cycle-consistent
  adversarial networks.
\newblock In {\em Proceedings of the IEEE international conference on computer
  vision}, pages 2223--2232, 2017.

\bibitem{zhu2017deep}
Xiao~Xiang Zhu, Devis Tuia, Lichao Mou, Gui-Song Xia, Liangpei Zhang, Feng Xu,
  and Friedrich Fraundorfer.
\newblock Deep learning in remote sensing: A comprehensive review and list of
  resources.
\newblock {\em IEEE Geoscience and Remote Sensing Magazine}, 5(4):8--36, 2017.

\end{thebibliography}
}

%%%%%%%%%%%%%%%%%%%%%%%%%%%%%%%%%%%%%%%%%%%%%%%%%%%%%%%%%%%%%%%%%%%%%%%%%%%%%%%%%%%%%%%%%%

%%%%%%%%%%%%%%%% SUPPLEMENTARY MATERIAL %%%%%%%%%%%%%%%% 
\newpage
\onecolumn
% \pagebreak
\appendix
\section{Supplementary Material}\label{sec:supp}

A demo video and additional quantitative results are provided below.

\subsection{Demo video}\label{sec:demos}
We provide a demo video at \url{https://youtu.be/5N6JDZf3pLQ} of a grey car being attacked with a patch in the Side Street scene and a blue car being attacked with a patch in the Car Park scene. The colour of a bounding box indicates the objectness score: green is for a score of above 0.8, red is for a score of between 0.5 and 0.8, and grey is for a score of below 0.5.

% \begin{itemize}[leftmargin=1em,itemsep=2pt,parsep=0pt,topsep=2pt]
% 	\item sidestreet-grey-clean: targeted moving car with no patch tracked and detected, 
% 	\item sidestreet-grey-patch-dynamic: targeted moving car with Type ON patch tracked and detected,
% 	\item sidestreet-grey-patch-shade: targeted static car with Type ON patch in the shade tracked and detected,
% 	\item sidestreet-grey-patch-sun: targeted static car with Type ON patch in the sun tracked and detected,
% 	\item sidestreet-grey-patchoff-sun: target static car with Type OFF in the sun tracked and detected,
% 	\item carpark-all-clean: all cars detected in the scene of attack by the car detector,
% 	\item carpark-blue-clean: targeted static car with no patch in the shade tracked and detected,
% 	\item carpark-blue-patch-dynamic: targeted static car with Type ON patch in the shade tracked and detected.
% \end{itemize}

\subsection{Additional quantitative results}\label{sec:additional_results}

% Below, Sec.~4.1 = Sec.~\ref{sec:yolov3}, Sec.~4.3 = Sec.~\ref{sec:path}, Table~1 =  Table~\ref{tab:aorr_results}

\begin{figure*}[ht]
     \centering
     \begin{subfigure}[b]{0.45\textwidth}
         \centering
         \includegraphics[width=\textwidth]{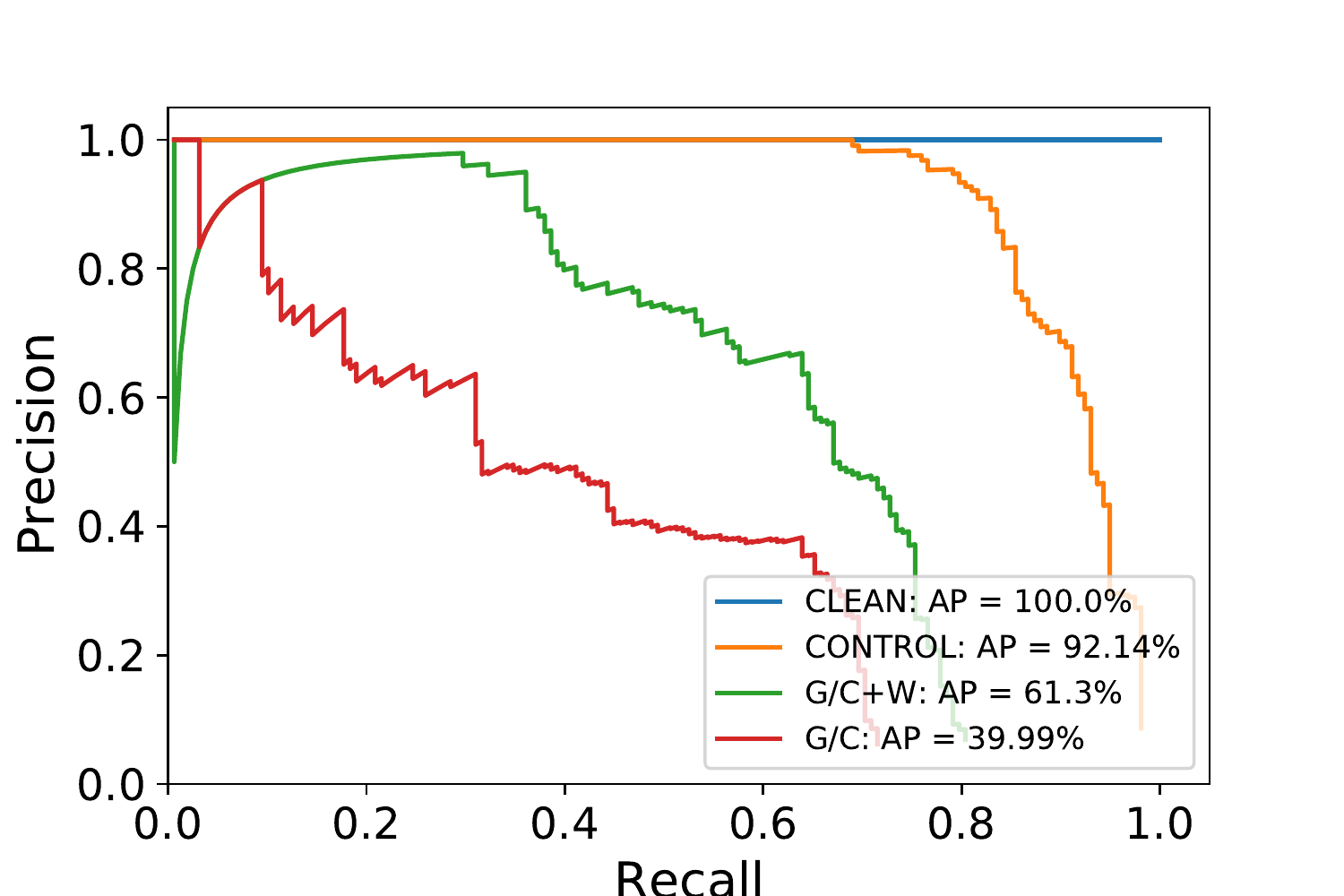}
         \caption{Type ON patch - Side Street (STD)}
         \label{fig:ap1}
     \end{subfigure}
     \hfill
     \begin{subfigure}[b]{0.45\textwidth}
         \centering
         \includegraphics[width=\textwidth]{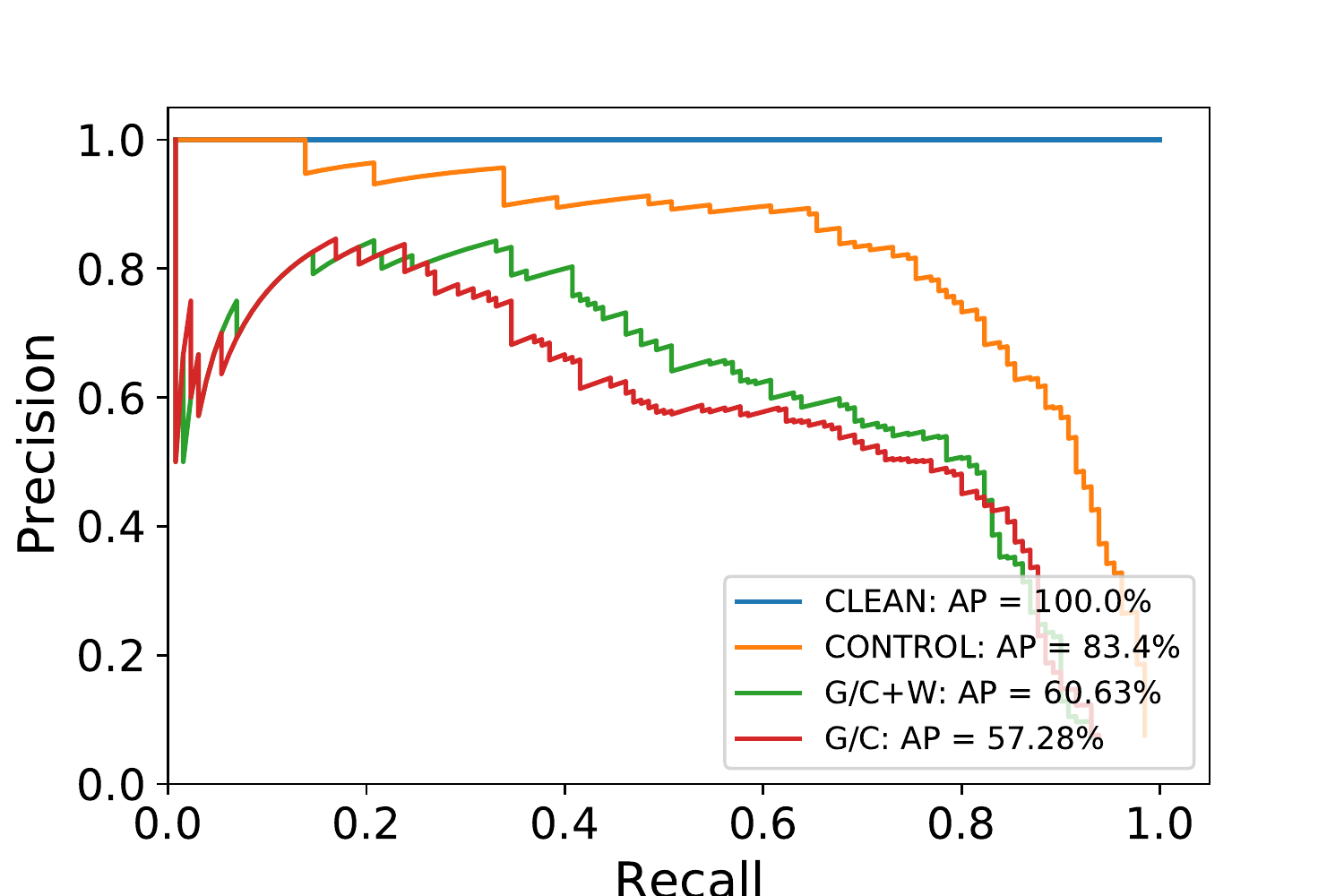}
         \caption{Type ON patch - Side Street (STD-W)}
         \label{fig:ap2}
     \end{subfigure}
     \hfill
     \begin{subfigure}[b]{0.45\textwidth}
         \centering
         \includegraphics[width=\textwidth]{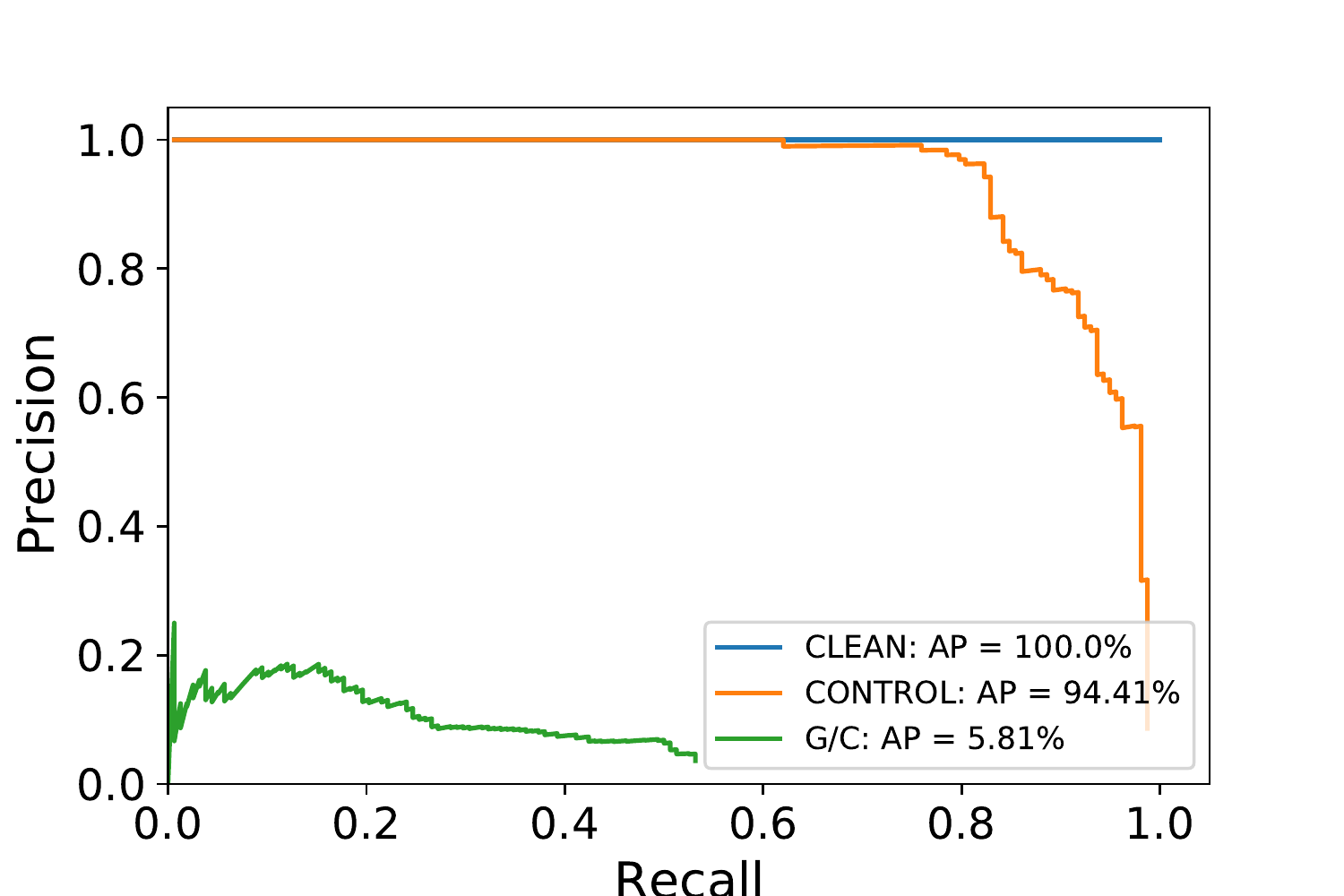}
         \caption{Type OFF patch - Side Street (STD)}
         \label{fig:ap3}
     \end{subfigure}
     \hfill
     \begin{subfigure}[b]{0.45\textwidth}
         \centering
         \includegraphics[width=\textwidth]{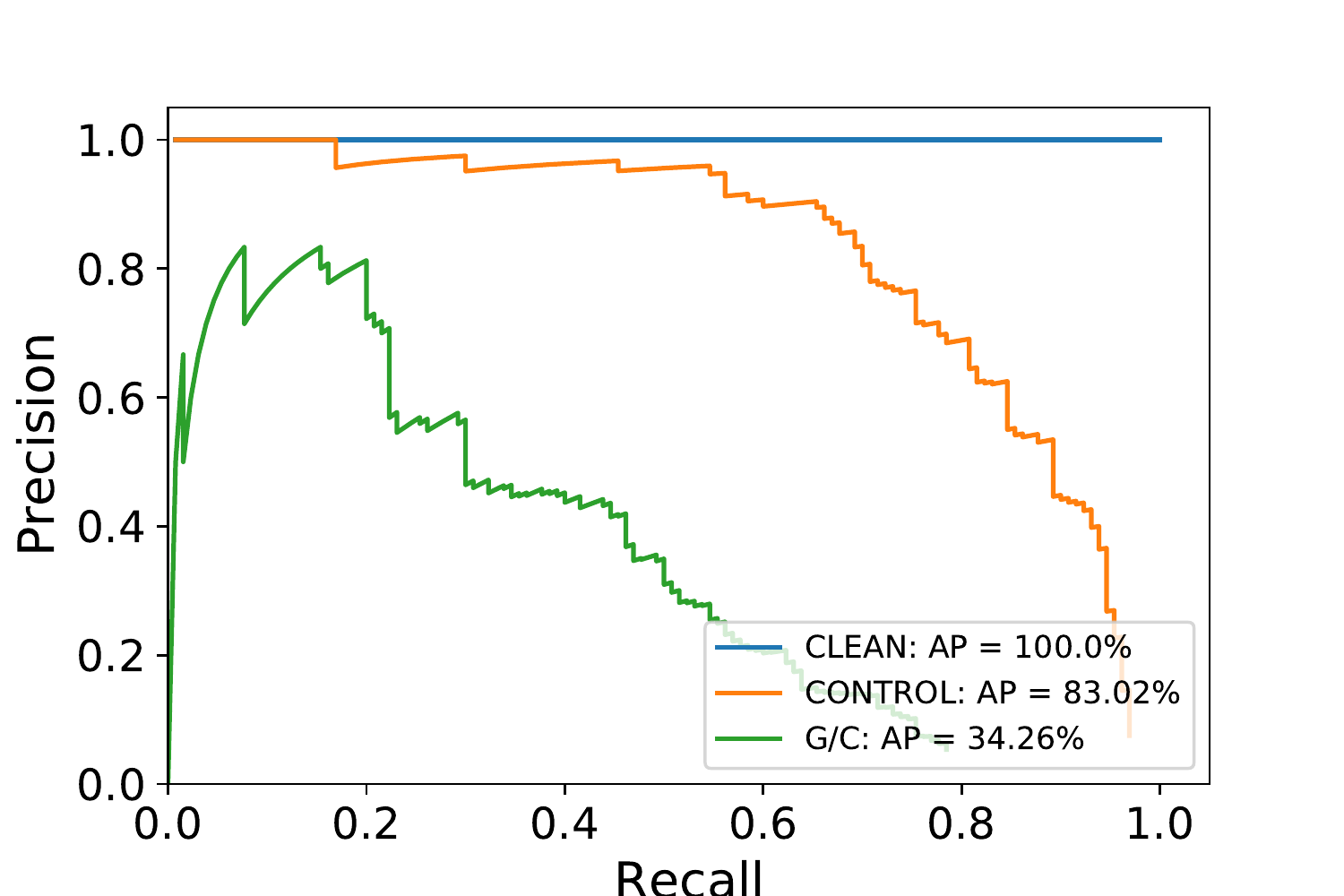}
         \caption{Type OFF patch - Side Street (STD-W)}
         \label{fig:ap4}
     \end{subfigure}
     \hfill
     \begin{subfigure}[b]{0.45\textwidth}
         \centering
         \includegraphics[width=\textwidth]{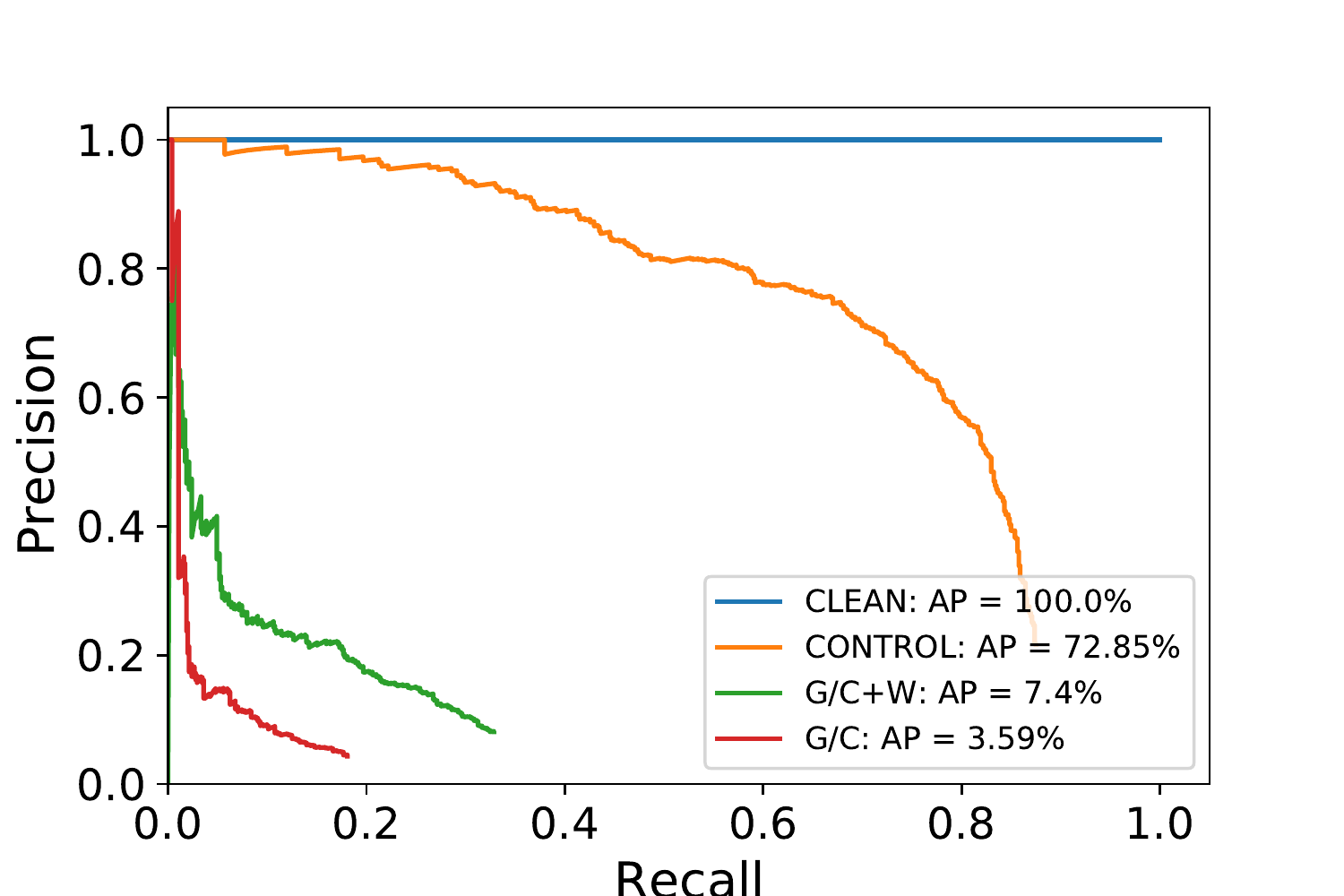}
         \caption{Type ON patch - Car Park (STD)}
         \label{fig:ap5}
     \end{subfigure}
     \hfill
     \begin{subfigure}[b]{0.45\textwidth}
         \centering
         \includegraphics[width=\textwidth]{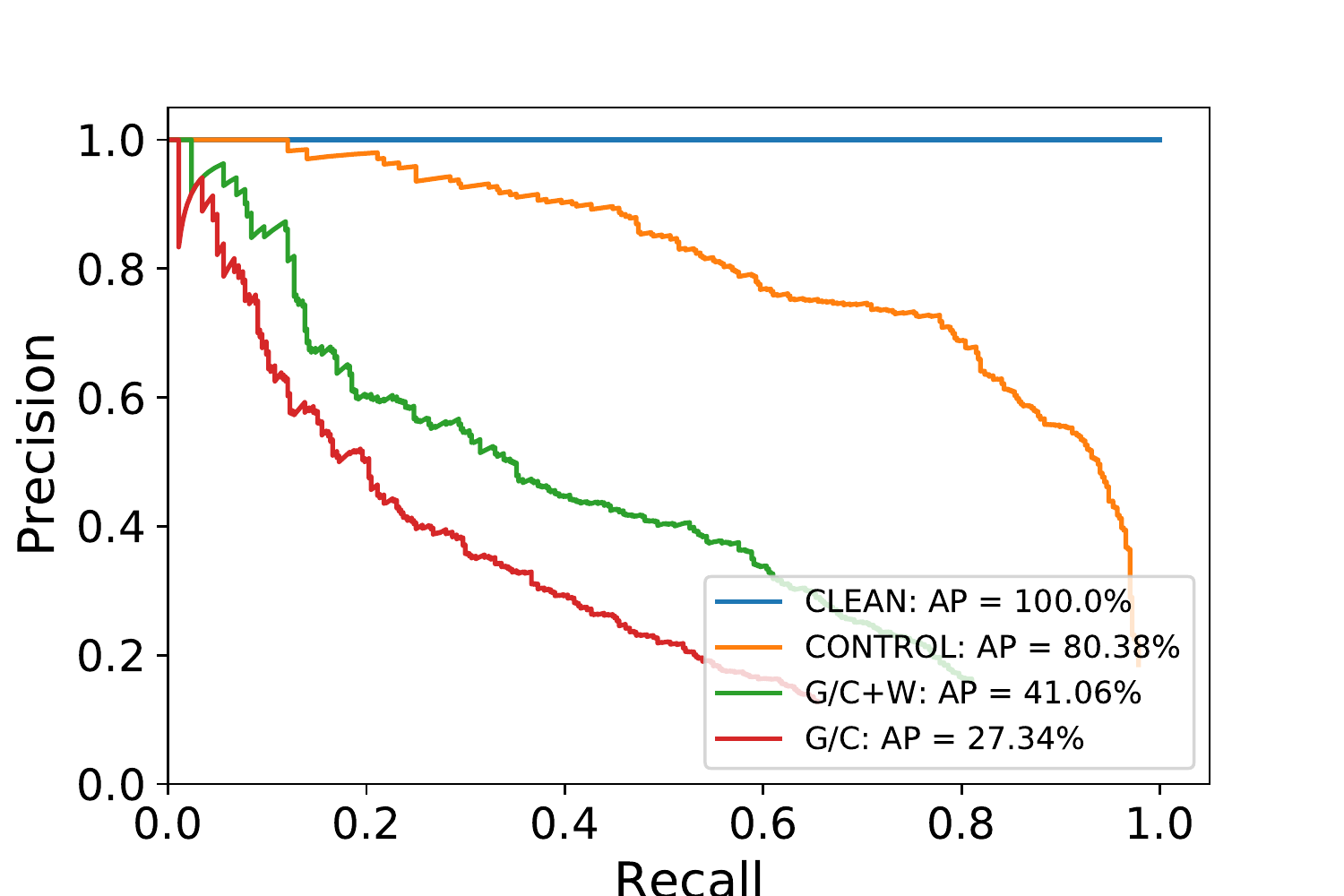}
         \caption{Type ON patch - Car Park (STD-W)}
         \label{fig:ap6}
     \end{subfigure}
        \caption{The precision-recall curves as well as AP values of the car detector for different variants of the pipeline and testing regimes.}
        \label{fig:ap}
\end{figure*}

Since we were not interested in misclassifying cars, we only provide the precision-recall curves and average precision (AP) values (see Fig.~\ref{fig:ap}) of the car detector (Sec.~4.1) when it is attacked with patches optimised under different variants of the pipeline (Sec.~4.3). The CLEAN curve is the evaluation of testing images with no patches applied. The CONTROL curve is with patches with random intensity patterns applied. The G/C+W curve is with patches optimised with geometric, colour-space and weather augmentations applied (full pipeline), and the G/C curve is with patches optimised with only geometric and colour-space augmentations. These patches were also evaluated under two testing regimes. STD are testing images with no weather effects applied and STD-W is with weather effects applied. 

Similar to the results in Table~1, Fig.~\ref{fig:ap} shows that, while both G/C and G/C+W were significantly more effective (lower AP) than Control, G/C visibly outperforms G/C+W. Again, this indicated the lack of value in performing weather augmentations during training and further motivated us to ignore G/C+W for training Type OFF patches for Side Street. Type OFF G/C patches were also found to outperformed Type ON G/C patches for the same scene (Side Street).

\end{document}